\documentclass{article}

\PassOptionsToPackage{numbers,sort&compress}{natbib}

\usepackage[preprint]{neurips_2026}

\usepackage[utf8]{inputenc}
\usepackage[T1]{fontenc}
\usepackage[hidelinks]{hyperref}
\usepackage{url}
\usepackage{booktabs}
\usepackage{amsfonts}
\usepackage{amsthm}
\usepackage{nicefrac}
\usepackage{microtype}
\usepackage{xcolor}
\usepackage{fancyvrb}
\definecolor{codekeyword}{HTML}{553A99}
\definecolor{codecomment}{HTML}{356B46}
\definecolor{codestring}{HTML}{9B3535}
\definecolor{codefunction}{HTML}{215B8C}
\definecolor{codebackground}{HTML}{F6F7F9}
\definecolor{codeborder}{HTML}{D8DCE3}

\newcommand{\todo}[1]{}
\usepackage{graphicx}
\usepackage{amsmath}
\usepackage{pgfplots}
\usepackage{algorithm}
\usepackage{algpseudocode}
\usepackage{makecell}
\usepackage{subcaption}

\pgfplotsset{compat=1.18}

\title{TreemapMix: Dirichlet-Controlled Multi-Image Augmentation for Probability and Ordinal Supervision}

\author{%
  \begin{tabular}[t]{@{}cc@{}}
    \textbf{Ejafa Bassam}\thanks{Equal contribution.}\hspace{0.6em}\thanks{This work was conducted while Ejafa Bassam was at Peking University.} &
    \textbf{Konstantin Garbers}\textsuperscript{1}\footnotemark[1]\\
    \texttt{ejafabassam@alumni.pku.edu.cn} & \texttt{konstantin.garbers25@stu.pku.edu.cn}\\[0.5ex]
    \textbf{Yingsheng Geng}\textsuperscript{2} &
    \textbf{Dalton Jens}\textsuperscript{1}\\
    \texttt{gys@bupt.edu.cn} & \texttt{daltonjens@gmail.com}\\[0.5ex]
    \textbf{Jiaxu Qian}\textsuperscript{1} &
    \textbf{Guyue Liu}\textsuperscript{1}\thanks{Corresponding author.}\\
    \texttt{jiaxu@pku.edu.cn} & \texttt{guyue@pku.edu.cn}
  \end{tabular}\\[0.5ex]
  \textsuperscript{1}Peking University\\
  \textsuperscript{2}Beijing University of Posts and Telecommunications
}

\begin{document}

\maketitle

\begin{abstract}
Data augmentation improves visual recognition by exposing models to synthetic training examples that encourage generalization beyond the observed data. Mixed-sample methods extend this idea by combining multiple images. However, many common approaches either mix only pairs of images or rely on fixed small layouts, limiting control over source count, visible area proportions, and induced supervision. We propose TreemapMix, a distribution-controlled multi-image augmentation that samples source weights from a Dirichlet distribution and assigns them to area-proportional regions using a treemap partition. The resulting known region areas support two complementary objectives. TreemapMix with soft cross-entropy (SCE) treats areas as soft class-probability targets, while TreemapMix with a Plackett-Luce (PL) objective converts areas into ordinal supervision. On ImageNet-1K under matched training recipes, TreemapMix-SCE achieves substantially lower calibration error than evaluated mixed-sample baselines, while TreemapMix-PL achieves the strongest Top-1 accuracy. Transfer experiments on object detection and instance segmentation suggest that TreemapMix-pretrained features remain useful beyond classification. These results show that controlled multi-image composition can expose the same region-area information as either probability-matching or rank-matching supervision.
Code is available at \href{https://github.com/Ejafa/treemapmix}{\textcolor{codefunction}{github.com/Ejafa/treemapmix}}.
\end{abstract}

\section{Introduction}

Image classification is usually formulated as learning one target label from one
image. Natural visual experience is more compositional: multiple objects often
appear in the same field of view, and visual categories are often understood
through comparisons among related sources of evidence. This suggests a
complementary augmentation principle: construct training examples that expose
several labeled sources jointly, while preserving explicit control over how much
visual evidence each source contributes.

\begin{figure*}[t]
  \centering
  \setlength{\tabcolsep}{2pt}
  \begin{tabular}{ccccc}
    \includegraphics{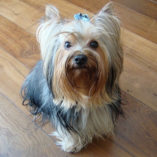} &
    \includegraphics{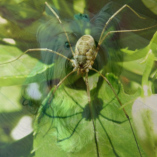} &
    \includegraphics{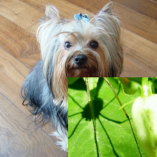} &
    \includegraphics{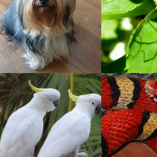} &
    \includegraphics{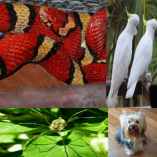} \\
    (a) None & (b) Mixup & (c) CutMix & (d) RICAP & (e) TreemapMix
  \end{tabular}
  \caption{Qualitative comparison of five augmentation strategies on matched source images using resize-and-center-crop preprocessing only. Mixup blends full images, CutMix pastes a random region, RICAP composes four tiles, and TreemapMix composes $K$ rescaled source images within a randomized layout.}
  \label{fig:aug-showcase-clean}
\end{figure*}

Mixed-sample data augmentation moves in this direction by composing examples
and mixing their labels. Mixup~\citep{mixup} and CutMix~\citep{cutmix} are
widely used in modern image-classification recipes~\citep{
wightmanResNetStrikesBack2021,pmlr-v139-touvron21a,
touvronDeiTIIIRevenge2022,liOpenMixupOpenMixup2024}, but they are primarily
pairwise: Mixup globally interpolates two images, and CutMix replaces one
region with a patch from another image. Multi-image methods such as
RICAP~\citep{takahashiDataAugmentationUsing2020} and Mosaic-style
augmentation~\citep{bochkovskiyYOLOv4OptimalSpeed2020} show the value of
composing several images, but typically use a fixed small number of sources and
restricted layout families. Moreover, existing methods usually treat region
area only as a soft-label weight, rather than as a controllable object that can
shape both the visual composition and the supervision signal. Figure~\ref{fig:aug-showcase-clean}
shows a qualitative comparison of these augmentation patterns on matched source
images.

We propose \emph{TreemapMix}, a $K$-way image-mixing augmentation that composes
multiple single-label images into a shared canvas while preserving the training
sample count. TreemapMix samples region weights from a Dirichlet distribution
and maps them to non-overlapping area-proportional regions using a squarified
treemap layout~\citep{squarifiedtreemap}. The Dirichlet concentration
parameter $\alpha$ controls the composition, ranging from dominant-region
mixtures to more balanced multi-image mixtures. To prevent persistent
label-to-area or label-to-rank bias, TreemapMix generates $K$ cyclic views from
each group of $K$ source images, so that every source image is paired with every
sampled region weight across the cyclic orbit. Area-preserving symmetries then
increase spatial diversity without changing the target weights.

TreemapMix also exposes a supervision choice. Region area provides a natural
but imperfect proxy for visual evidence: larger regions usually contain more
pixels, more context, and more object parts, and are therefore often easier to
recognize than smaller regions. This suggests an ordinal learning signal within
a composed image, where labels assigned to larger regions should generally
receive higher scores than labels assigned to smaller regions. However, area is
not a calibrated measure of semantic evidence; a large region may contain mostly
background, while a small region may contain a discriminative object part. We
therefore study two complementary objectives: TreemapMix-SCE, which treats
region areas as soft probability targets and optimizes soft cross-entropy, and
TreemapMix-PL, which converts region areas into area-induced ordinal
supervision by ranking the target classes and optimizing a Plackett-Luce
listwise objective~\citep{luce1959,plackett1975,listmle}.

We evaluate TreemapMix on ImageNet-1K~\citep{
dengImageNetLargescaleHierarchical2009} under matched training recipes across
multiple modern image-classification backbones. Compared with 13 baseline configurations, TreemapMix-PL achieves
the strongest Top-1 accuracy across the evaluated ImageNet-1K backbones, while
TreemapMix-SCE gives the lowest expected calibration error (ECE). The original
NLL comparison also favors SCE, while Brier-score gains are smaller and mixed
(Appendix~\ref{app:in1k-confidence-results}). Further
diagnostics show that TreemapMix-PL improves alignment between logits and the
area-induced ordering of composed regions, and COCO transfer experiments
indicate that TreemapMix-pretrained representations remain useful for detection
and instance segmentation~\citep{linMicrosoftCOCOCommon2015}.

Our contributions are:
\begin{enumerate}
    \setlength{\itemsep}{1pt}
    \setlength{\parsep}{0pt}
    \setlength{\topsep}{2pt}
    \renewcommand{\labelenumi}{\roman{enumi}.}
    \item We introduce TreemapMix, a Dirichlet-controlled $K$-way image-mixing
    augmentation that supports variable-source composition with explicit
    area-proportional treemap layouts while preserving the training sample
    count.

    \item We formulate two supervision objectives from the same composed image:
    area-proportional soft cross-entropy for probability matching and
    Plackett-Luce ranking for area-induced ordinal supervision.

    \item We analyze how the Dirichlet concentration parameter controls target
    variability, probability-matching supervision, and ranking emphasis.

    \item We provide an empirical evaluation on ImageNet-1K, including
    accuracy, calibration metrics, area-logit alignment analysis, and COCO
    detection and instance-segmentation transfer experiments. We also evaluate
    long-horizon training, loader throughput and host memory, long-tailed
    recognition, and additional model architectures.

\end{enumerate}

\section{Related Work}

Standard augmentations preserve the original label, including region dropout
methods such as Cutout and Random Erasing~\citep{devries2017cutout,
zhong2020random} and policy-based methods such as AutoAugment, RandAugment,
and AugMix~\citep{cubuk2019autoaugment,
cubukRandAugmentPracticalAutomated2019,hendrycks2020augmix}. Pairwise
mixed-sample methods instead combine two examples and assign mixed targets:
Mixup~\citep{mixup} interpolates images and labels, while CutMix~\citep{cutmix}
uses spatial replacement with area-based labels. Later methods refine pairwise
composition or label estimation through resized patches, Fourier masks, grids,
saliency, class activation maps, attention, or token-level mixing~\citep{
qin2020resizemixmixingdatapreserved,harris2021fmixenhancingmixedsample,
baekGridMixStrongRegularization2021,kim2020puzzlemixexploitingsaliency,
huang2020snapmixsemanticallyproportionalmixing,
chen2021transmixattendmixvision,liu2023tokenmixrethinkingimagemixing}.
These methods improve label-preserving or two-source augmentation, whereas
TreemapMix constructs multi-source examples with a controllable distribution
over two or more visible labels.

\paragraph{Multi-image augmentation and label supervision:}
Multi-image augmentations compose more than two sources, including RICAP, which
patches four crops with area-weighted labels~\citep{
takahashiDataAugmentationUsing2020}, Mosaic augmentation in detection
pipelines~\citep{bochkovskiyYOLOv4OptimalSpeed2020}, Cut-thumbnail
~\citep{Xie_2021}, StackMix~\citep{chen2022stackmix}, and
DCutMix~\citep{jeong2022dcutmix}. Other methods generalize mixing beyond
spatial layouts, such as k-Mixup through optimal-transport interpolation
~\citep{greenewald2023kmixupregularizationdeeplearning} and MixMo through a
multi-input multi-output architecture~\citep{rame2021mixmo}. TreemapMix is
closest to spatial multi-image methods such as RICAP and Mosaic, but samples a
single coherent $K$-way area distribution from a Dirichlet distribution and
realizes it as an area-proportional treemap layout; unlike k-Mixup and MixMo,
it remains spatially explicit and uses a standard single-input model interface.
Most spatial mixing methods treat visible area as a soft-label
weight, although area may be an imperfect proxy for semantic evidence. Related
work studies confidence and target behavior under mixup-style training through
SmoothMix for randomized smoothing~\citep{jeong2021smoothmixtrainingconfidencecalibratedsmoothed},
RegMixup~\citep{pinto2023regmixupmixupregularizersurprisingly}, and
RankMixup~\citep{noh2023rankmixuprankingbasedmixuptraining}. Plackett-Luce
ranking has also been used for classification in Plackett--Luce Distillation (PLD)~\citep{pld}, but in a
knowledge-distillation setting where a teacher model provides the class ranking.
In contrast, TreemapMix induces the ranking directly from the mixed input: the
Dirichlet-sampled region areas define the area-induced ordering over target
classes. TreemapMix-SCE follows the area-weighted probability interpretation,
while TreemapMix-PL converts the same areas into ordinal supervision with a
Plackett-Luce ranking objective.

\section{Method}\label{sec:method}

TreemapMix is a $K$-way image-mixing augmentation that replaces each group of $K$ source images with $K$ cyclic mixed samples. Each mixed sample is supervised either by a soft probability-matching objective or by a ranking-based Plackett--Luce objective. We use integer $K\ge1$ for the group size, $C$ for the number of classes, and \(\mathcal{G}=\{(x_k,y_k)\}_{k=1}^{K}\), where \(y_k\in\{1,\dots,C\}\), for a source group.

\subsection{TreemapMix Augmentation}\label{sec:labelmix}

For each group, sample $w\sim\operatorname{Dirichlet}(\alpha,\ldots,\alpha)$, $\alpha>0$, with $\sum_k w_k=1$. These are slot weights, not fixed source-image weights. Smaller $\alpha$ favors dominant-region mixtures; larger $\alpha$ yields more balanced mixtures.

For the balanced-data protocol, TreemapMix first forms class-balanced rows from the training set and then splits each row into groups of size $K$. This encourages diverse multi-class mixtures and reduces the tendency of frequent classes to dominate mixed samples. When $K\le C$, complete groups formed from a class-balanced row contain distinct labels by construction.

For stable layout construction and for ranking-based supervision, TreemapMix sorts the sampled weights in ascending order: \refstepcounter{equation}\label{eq:ascending-weights}\(\lambda_1\le\lambda_2\le\cdots\le\lambda_K,\lambda_i=w_{\tau(i)}\)\nobreakspace\textup{(\theequation)}.
Here $\tau$ is a permutation of the source indices. The sorted weights define $K$ layout slots. Slot $i$ has area weight $\lambda_i$, with slot $1$ the smallest-area slot and slot $K$ the largest-area slot.
Given the sorted weights, TreemapMix partitions the image canvas $\Omega=[0,W)\times[0,H)$ into non-overlapping rectangular boxes \(\mathcal{B}=\{B_1,\dots,B_K\}\) with \(|B_i|/(HW)\approx \lambda_i\), such that \(B_i\cap B_j=\emptyset\) for \(i\neq j\) and \(\bigcup_{i=1}^{K}B_i=\Omega\).

Each box includes its left/top boundaries and excludes its right/bottom boundaries, so adjacent boxes share no pixels. We use a squarified treemap layout~\cite{squarifiedtreemap} to obtain area-proportional regions while favoring compact rectangles. This helps preserve useful semantic content inside each patch. The normalized treemap boxes are converted to integer pixel coordinates. If integer conversion yields zero-area boxes, TreemapMix falls back to a stripe layout with approximate area matching; target weights remain unchanged.

Area-preserving spatial symmetries increase diversity without changing targets: the eight $D_4$ symmetries for square images, or identity, $180^\circ$ rotation, and horizontal/vertical flips for rectangular images.

TreemapMix then generates $K$ cyclic views from each group. Let slot $i$ have weight $\lambda_i$, box $B_i$, and binary mask $M_i$. For cyclic view $s\in\{0,\dots,K-1\}$, define \refstepcounter{equation}\label{eq:cyclic-map}\(\pi_s(i)=1+\bigl((i+s-1)\bmod K\bigr)\)\nobreakspace\textup{(\theequation)}. The $s$-th mixed image is \(\tilde{x}^{(s)}=\sum_{i=1}^{K} M_i\odot \operatorname{Resize}(x_{\pi_s(i)},B_i)\), with sparse label-weight target \refstepcounter{equation}\label{eq:cyclic-target}\(\mathcal{T}^{(s)}=\{(y_{\pi_s(i)},\lambda_i)\}_{i=1}^{K}\)\nobreakspace\textup{(\theequation)}.
Here $\odot$ denotes elementwise multiplication, and $\operatorname{Resize}$ embeds the resized patch in the full canvas. Thus, in each cyclic view, the source image assigned to slot $i$ receives the slot weight $\lambda_i$ as its target weight.

All $K$ cyclic views are retained for complete groups, avoiding enumeration of $K!$ assignments. Each source visits each slot once, so mean SCE target mass is $m_c/K$ for source-label multiplicity $m_c$, uniform for distinct labels (Lemma~\ref{lem:cyclic-balancing}). This prevents persistent source-to-slot bias, but does not remove prediction-dependent PL competition. Partial orbits lack the full balancing guarantee.

Appendix~\ref{app:labelmix-details} gives pseudocode, remainder handling, geometry, and fallback details. Output count matches the constructed row stream. The sparse target supports soft probability supervision, or PL ranking for distinct-label groups.

\subsection{Soft TreemapMix Cross-Entropy}\label{sec:soft-ce}

Let $z=f_\theta(\tilde{x})$ be the model logits (class scores before softmax), with $p_c=\operatorname{softmax}(z)_c$. We write $\mathbf{1}[\cdot]$ for the indicator function. For a TreemapMix sample $\tilde{x}$ with sparse target \(\mathcal{T}=\{(c_i,\eta_i)\}_{i=1}^{K}\), where \(\eta_i\ge 0\) and \(\sum_{i=1}^{K}\eta_i=1\), and where $c_i$ is the class assigned to slot $i$ and $\eta_i$ is its area weight, TreemapMix defines \refstepcounter{equation}\label{eq:soft-target}\(q_c=\sum_{i=1}^{K}\eta_i\mathbf{1}[c_i=c], p_c=p_\theta(c\mid\tilde{x})\)\nobreakspace\textup{(\theequation)}.
For cyclic view $s$, we have $c_i=y_{\pi_s(i)}$ and $\eta_i=\lambda_i$. For the distinct-label groups used in the balanced-data protocol, $q_{c_i}=\eta_i$. Thus $q$ is a probability distribution over the $C$ classes.
We denote the cross-entropy objective of TreemapMix-SCE and the ranking objective of TreemapMix-PL by $\mathcal{L}_{\mathrm{TM\text{-}CE}}$ and $\mathcal{L}_{\mathrm{TM\text{-}PL}}$, respectively. We obtain the soft cross-entropy objective by minimizing the KL divergence~\cite{kullback1951} between $q$ and $p_\theta$. Since the entropy of $q$ is independent of the model parameters, this is equivalent to minimizing
\begin{equation}
\label{eq:lm-ce}
\mathcal{L}_{\mathrm{TM\text{-}CE}}=-\sum_{c=1}^{C}q_c\log p_\theta(c\mid\tilde{x})=-\sum_{i=1}^{K}\eta_i\log p_\theta(c_i\mid\tilde{x}).
\end{equation}
For a mini-batch of size $B$, the average loss is \refstepcounter{equation}\label{eq:lm-ce-batch}\(\mathcal{L}_{\mathrm{TM\text{-}CE}}=-\frac{1}{B}\sum_{b=1}^{B}\sum_{i=1}^{K}\eta_{b,i}\log p_\theta(c_{b,i}\mid\tilde{x}_b)\)\nobreakspace\textup{(\theequation)}.
Thus, TreemapMix-SCE treats the TreemapMix weights as area-proportional probability targets and encourages predicted probability mass to match the mixed-image area weights.

The per-sample logit gradient has the standard prediction-minus-target form: \refstepcounter{equation}\label{eq:lm-ce-grad}\(\frac{\partial \mathcal{L}_{\mathrm{TM\text{-}CE}}}{\partial z_c}=p_c-q_c\)\nobreakspace\textup{(\theequation)}.
For the batch mean, the derivative with respect to each sample's logits additionally includes $1/B$.
For complete distinct-label groups, the full cyclic orbit has exactly balanced mean targets, while a uniformly selected cyclic view retains Dirichlet variability. Appendix Lemma~\ref{lem:ce-cyclic-variance} derives its $\alpha$-controlled variance under unfiltered sampling. Fixed rank positions instead follow Dirichlet order statistics. With logits fixed, the gradient $p_c-q_c$ has the same target-side variance.

\subsection{TreemapMix Plackett--Luce Loss}\label{sec:pl}

As motivated in the introduction, area is an imperfect proxy for semantic evidence. The PL objective uses the same weights as an ordinal signal, encouraging larger-area target classes to rank above smaller-area classes rather than matching their probabilities exactly.

The PL objective assumes distinct target labels. Let $\mathcal{T}=\{(c_i,\eta_i)\}_{i=1}^{K}$ be a target with distinct $c_i$, and define a slot permutation $\sigma$ with $r_\ell=c_{\sigma(\ell)}$ and class-associated weight $\rho_{r_\ell}=\eta_{\sigma(\ell)}$. Order these weights descending: \refstepcounter{equation}\label{eq:descending-ranking}\(\rho_{r_1}\ge \rho_{r_2}\ge \cdots \ge \rho_{r_K}, r_1\succ r_2\succ\cdots\succ r_K\)\nobreakspace\textup{(\theequation)}. Let $\mathcal{U}=\{1,\dots,C\}\setminus\{r_1,\dots,r_K\}$ be the set of classes absent from the TreemapMix target, and let $p_{\mathcal U}=1-\sum_{j=1}^{K}p_{r_j}$ be their total predicted probability mass.
The ranking likelihood selects classes sequentially from largest to smallest target weight, normalizing each choice over the remaining classes. Following the Plackett--Luce model~\cite{luce1959,plackett1975} and listwise maximum-likelihood (ListMLE) objectives~\cite{listmle}, the descending TreemapMix-PL loss is
\begin{equation}
\label{eq:lm-pl-desc}
\mathcal{L}_{\mathrm{TM\text{-}PL}}=-\sum_{\ell=1}^{K}\rho_{r_\ell}\log \frac{p_\theta(r_\ell\mid\tilde{x})}{p_{\mathcal U}+\sum_{j=\ell}^{K}p_\theta(r_j\mid\tilde{x})}
\end{equation}
Similar to position-aware listwise objectives such as Position-Aware ListMLE~\cite{plistmle} and PLD~\cite{pld}, TreemapMix-PL weights each ranking term by its target importance; this weighted surrogate is not itself a normalized ranking likelihood. TreemapMix-SCE and TreemapMix-PL therefore use the same TreemapMix target in complementary ways: TreemapMix-SCE seeks \(p_\theta(c\mid\tilde{x})\approx q_c\), whereas TreemapMix-PL seeks \(r_1\succ r_2\succ\cdots\succ r_K\).

For implementation, we use the ascending order. Let \(a_i=r_{K+1-i}\), \(\gamma_i=\rho_{a_i}\), and \(\gamma_1\le\gamma_2\le\cdots\le\gamma_K\). Thus, $a_1$ is the weakest ranked target class and $a_K$ is the strongest. Define \refstepcounter{equation}\label{eq:Di}\(D_i=p_{\mathcal U}+\sum_{j=1}^{i}p_{a_j}=1-\sum_{j=i+1}^{K}p_{a_j}\)\nobreakspace\textup{(\theequation)}. Then the loss becomes \refstepcounter{equation}\label{eq:lm-pl-asc}\(\mathcal{L}_{\mathrm{TM\text{-}PL}}=\sum_{i=1}^{K}\gamma_i\left[\log D_i-\log p_{a_i}\right]\)\nobreakspace\textup{(\theequation)}. If \refstepcounter{equation}\label{eq:Ti}\(T_i=\{a_{i+1},\dots,a_K\}\)\nobreakspace\textup{(\theequation)} denotes the set of classes stronger than $a_i$, then the logit gradient is
\begin{equation}
\label{eq:lm-pl-grad}
\frac{\partial \mathcal{L}_{\mathrm{TM\text{-}PL}}}{\partial z_c}=\sum_{i=1}^{K}\gamma_i\left[\frac{p_c\bigl(1-\mathbf{1}[c\in T_i]\bigr)}{D_i}-\mathbf{1}[c=a_i]\right]
\end{equation}
At the logit level, this gradient suppresses unranked classes, promotes the strongest ranked class, and regulates intermediate ranked classes through competition with stronger classes. The full likelihood derivation, ascending reindexing, and gradient proof are given in Appendix~\ref{app:pl}.

For distinct target labels, the ascending PL weights satisfy \(\gamma_i=\lambda_i\). The sorted Dirichlet weights determine the importance assigned to each rank position. Appendix Lemma~\ref{lem:pl-order} in Appendix~\ref{app:pl} characterizes how \(\alpha\) controls this ranking emphasis through the order statistics of the sampled Dirichlet weights. This analysis holds logits and rank assignments fixed and assumes unfiltered Dirichlet sampling; filtering changes the order-statistic moments. As $\alpha$ increases, the sorted weights concentrate near $1/K$, so in the distinct-label case the loss approaches $1/K$ times an unweighted top-$K$ ListMLE-style objective. Equal-weight ties retain source-slot order.

For completeness, we also evaluate a convex combination of TreemapMix-SCE and
TreemapMix-PL. Let $\beta\in[0,1]$ denote the interpolation coefficient, where
$\beta=1$ recovers TreemapMix-SCE and $\beta=0$ recovers TreemapMix-PL. We
define \refstepcounter{equation}\label{eq:lm-mix}\(\mathcal{L}_{\mathrm{TM\text{-}Mix}}=\beta\mathcal{L}_{\mathrm{TM\text{-}CE}}+(1-\beta)\mathcal{L}_{\mathrm{TM\text{-}PL}}\)\nobreakspace\textup{(\theequation)}.
Using the ascending PL notation from Section~\ref{sec:pl}, this can be
written as \refstepcounter{equation}\label{eq:lm-mix-asc}\(\mathcal{L}_{\mathrm{TM\text{-}Mix}}=\sum_{i=1}^{K}\gamma_i\left[(1-\beta)\log D_i-\log p_{a_i}\right]\)\nobreakspace\textup{(\theequation)}.
This objective is not intended as a separate contribution; we
include it to test whether combining probability matching and rank matching
improves the accuracy--calibration trade-off.

\subsection{Dirichlet-Controlled Supervision}
\label{sec:gradient-supervision}

TreemapMix supports a fixed Dirichlet concentration or a schedule over training progress. We vary $\alpha$ between endpoints $\alpha_{\min}$ and $\alpha_{\max}$ using a linear or cosine schedule. A forward schedule increases $\alpha$, shifting from dominant-region mixtures toward more balanced mixtures; a backward schedule decreases it.

\section{Empirical Evaluation}
\paragraph{Protocol.}
Our main ImageNet-1K study uses four compact ViT/EVA02-family backbones in \texttt{timm}~\citep{fangEVA02VisualRepresentation2024}, two NVIDIA H20 GPUs per run, global batch size 1024, and a matched 110-epoch recipe. Unless specified otherwise, results are mean $\pm$ one sample standard deviation over seeds 42, 43, and 44. Accuracy is Top-1 percent; ECE is the sample-weighted absolute accuracy--confidence gap across 15 equal-width bins, in percentage points. TreemapMix uses $K=4$ for SCE/mixed, $K=6$ for PL, and cosine schedule $\alpha:0.1\to0.5$; mixed uses $\beta=0.5$. Hyperparameter settings selected on ViT-Wee are fixed across backbones. Appendix~\ref{app:appendix-aug-configs} gives compute, recipes, and hyperparameters; Appendix~\ref{app:loss-code} gives mathematical loss references.

\begin{figure*}[t]
\centering
\includegraphics{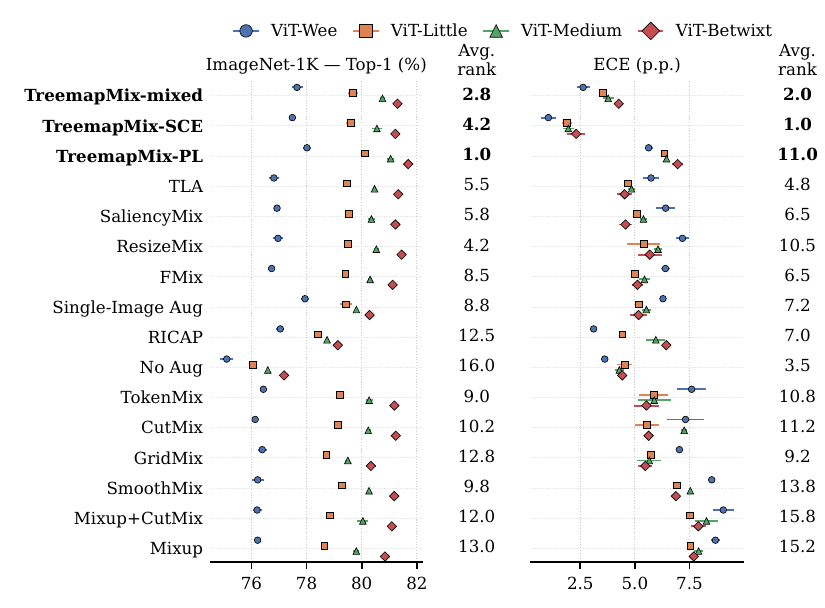}
\caption{ImageNet-1K accuracy and calibration for all 16 configurations under the matched 110-epoch protocol. Points show means over three seeds; horizontal intervals denote $\pm 1$ sample standard deviation. Marker shape and color identify the backbone. ECE uses 15 bins and is reported in percentage points. Average ranks are computed within each backbone (lower is better).}
\label{fig:in1k-main-results-top1}
\end{figure*}

\paragraph{ImageNet-1K.}
Tables~\ref{tab:in1k-full-results-acc} and \ref{tab:in1k-full-results-ece} report all 16 configurations, including the additional review-time baselines. FMix, GridMix, ResizeMix, SaliencyMix, SmoothMix, TokenMix, and TLA are OpenMixup-based reimplementations~\citep{liOpenMixupOpenMixup2024}, each selected from eight configurations on ViT-Wee. PL has the highest mean accuracy on every backbone, whereas SCE has the lowest ECE. The strongest baseline varies: single-image augmentation on Wee, SaliencyMix on Little, and ResizeMix on Medium/Betwixt. The mixed loss does not consistently improve on both objectives. Figure~\ref{fig:in1k-main-results-top1} compares Top-1 accuracy and ECE for all 16 configurations; NLL/Brier results appear in Appendix~\ref{app:in1k-confidence-results}.

\begin{table*}[!htbp]
  \caption{ImageNet-1K Top-1 accuracy (\%) under the matched 110-epoch protocol. Mean $\pm$ sample standard deviation over three seeds; bold marks the best mean per backbone.}
  \label{tab:in1k-full-results-acc}
  \centering
  \footnotesize
  \setlength{\tabcolsep}{4pt}
  \begin{tabular}{@{}lcccc@{}}
    \toprule
    Method & ViT-Wee $\uparrow$ & ViT-Little $\uparrow$ & ViT-Medium $\uparrow$ & ViT-Betwixt $\uparrow$ \\
    \midrule
    No Aug & $75.10 \pm 0.24$ & $76.05 \pm 0.14$ & $76.59 \pm 0.02$ & $77.18 \pm 0.14$ \\
    Single-Image Aug & $77.94 \pm 0.15$ & $79.43 \pm 0.22$ & $79.80 \pm 0.04$ & $80.28 \pm 0.16$ \\
    Mixup+CutMix & $76.20 \pm 0.16$ & $78.84 \pm 0.05$ & $80.04 \pm 0.20$ & $81.09 \pm 0.13$ \\
    Mixup & $76.22 \pm 0.12$ & $78.64 \pm 0.10$ & $79.80 \pm 0.11$ & $80.84 \pm 0.09$ \\
    CutMix & $76.13 \pm 0.07$ & $79.12 \pm 0.07$ & $80.23 \pm 0.09$ & $81.23 \pm 0.12$ \\
    RICAP & $77.05 \pm 0.15$ & $78.41 \pm 0.14$ & $78.74 \pm 0.07$ & $79.13 \pm 0.11$ \\
    FMix & $76.73 \pm 0.10$ & $79.40 \pm 0.06$ & $80.30 \pm 0.07$ & $81.12 \pm 0.10$ \\
    GridMix & $76.39 \pm 0.17$ & $78.72 \pm 0.05$ & $79.50 \pm 0.09$ & $80.33 \pm 0.18$ \\
    ResizeMix & $76.96 \pm 0.17$ & $79.51 \pm 0.15$ & $80.53 \pm 0.09$ & $81.44 \pm 0.13$ \\
    SaliencyMix & $76.93 \pm 0.11$ & $79.53 \pm 0.07$ & $80.35 \pm 0.12$ & $81.22 \pm 0.06$ \\
    SmoothMix & $76.22 \pm 0.22$ & $79.27 \pm 0.09$ & $80.26 \pm 0.08$ & $81.18 \pm 0.05$ \\
    TokenMix & $76.43 \pm 0.10$ & $79.19 \pm 0.13$ & $80.27 \pm 0.13$ & $81.18 \pm 0.10$ \\
    TLA & $76.81 \pm 0.19$ & $79.47 \pm 0.16$ & $80.46 \pm 0.03$ & $81.32 \pm 0.13$ \\
    TreemapMix-SCE & $77.48 \pm 0.03$ & $79.60 \pm 0.16$ & $80.55 \pm 0.17$ & $81.22 \pm 0.15$ \\
    TreemapMix-mixed & $77.65 \pm 0.20$ & $79.68 \pm 0.17$ & $80.75 \pm 0.09$ & $81.30 \pm 0.05$ \\
    TreemapMix-PL & $\mathbf{78.01 \pm 0.13}$ & $\mathbf{80.12 \pm 0.04}$ & $\mathbf{81.05 \pm 0.14}$ & $\mathbf{81.69 \pm 0.16}$ \\
    \bottomrule
  \end{tabular}
\end{table*}

\begin{table*}[!htbp]
  \caption{ImageNet-1K ECE (15 bins, percentage points). Same runs as Table~\ref{tab:in1k-full-results-acc}; bold marks the lowest mean per backbone.}
  \label{tab:in1k-full-results-ece}
  \centering
  \footnotesize
  \setlength{\tabcolsep}{4pt}
  \begin{tabular}{@{}lcccc@{}}
    \toprule
    Method & ViT-Wee $\downarrow$ & ViT-Little $\downarrow$ & ViT-Medium $\downarrow$ & ViT-Betwixt $\downarrow$ \\
    \midrule
    No Aug & $3.62 \pm 0.13$ & $4.53 \pm 0.36$ & $4.29 \pm 0.20$ & $4.42 \pm 0.02$ \\
    Single-Image Aug & $6.29 \pm 0.13$ & $5.19 \pm 0.12$ & $5.53 \pm 0.21$ & $5.17 \pm 0.40$ \\
    Mixup+CutMix & $9.05 \pm 0.47$ & $7.53 \pm 0.10$ & $8.28 \pm 0.51$ & $7.91 \pm 0.35$ \\
    Mixup & $8.69 \pm 0.20$ & $7.54 \pm 0.15$ & $7.92 \pm 0.20$ & $7.70 \pm 0.11$ \\
    CutMix & $7.32 \pm 0.84$ & $5.56 \pm 0.54$ & $7.26 \pm 0.14$ & $5.63 \pm 0.20$ \\
    RICAP & $3.11 \pm 0.02$ & $4.43 \pm 0.11$ & $5.96 \pm 0.43$ & $6.44 \pm 0.11$ \\
    FMix & $6.40 \pm 0.19$ & $5.01 \pm 0.10$ & $5.44 \pm 0.26$ & $5.12 \pm 0.26$ \\
    GridMix & $7.04 \pm 0.15$ & $5.72 \pm 0.17$ & $5.66 \pm 0.54$ & $5.48 \pm 0.32$ \\
    ResizeMix & $7.18 \pm 0.28$ & $5.40 \pm 0.74$ & $6.06 \pm 0.17$ & $5.68 \pm 0.55$ \\
    SaliencyMix & $6.41 \pm 0.43$ & $5.09 \pm 0.09$ & $5.39 \pm 0.16$ & $4.57 \pm 0.29$ \\
    SmoothMix & $8.52 \pm 0.06$ & $6.93 \pm 0.18$ & $7.54 \pm 0.06$ & $6.88 \pm 0.21$ \\
    TokenMix & $7.60 \pm 0.65$ & $5.86 \pm 0.65$ & $5.89 \pm 0.75$ & $5.53 \pm 0.59$ \\
    TLA & $5.74 \pm 0.36$ & $4.68 \pm 0.03$ & $4.85 \pm 0.17$ & $4.53 \pm 0.36$ \\
    TreemapMix-SCE & $\mathbf{1.04 \pm 0.35}$ & $\mathbf{1.89 \pm 0.23}$ & $\mathbf{1.95 \pm 0.25}$ & $\mathbf{2.31 \pm 0.41}$ \\
    TreemapMix-mixed & $2.63 \pm 0.30$ & $3.54 \pm 0.05$ & $3.78 \pm 0.26$ & $4.26 \pm 0.13$ \\
    TreemapMix-PL & $5.63 \pm 0.18$ & $6.35 \pm 0.10$ & $6.45 \pm 0.05$ & $6.96 \pm 0.25$ \\
    \bottomrule
  \end{tabular}
\end{table*}

\paragraph{Other datasets and architectures.}
Appendix Tables~\ref{tab:cifar100-full-results-acc} and \ref{tab:cifar100-full-results-ece} contain the complete CIFAR-100 comparison at 256-pixel resolution under 210 epochs including 10 warmup epochs. SCE has the lowest ECE on all backbones; PL has the highest accuracy on three, while ResizeMix leads on Little. Under the main ImageNet-1K 110-epoch protocol and TreemapMix settings, Table~\ref{tab:architecture-results} retains the accuracy/calibration separation on ConvNeXtV2-Nano, DeiT-Tiny, and Swin-Tiny. Small PL--SCE accuracy gaps on DeiT/Swin are not claimed statistically significant.

\begin{table*}[htbp]
  \caption{Cross-architecture ImageNet-1K results under 110 epochs. Mean $\pm$ sample standard deviation over three seeds; accuracy in percent and ECE in percentage points (15 bins). Bold marks the best mean per metric.}
  \label{tab:architecture-results}
  \centering
  \footnotesize
  \setlength{\tabcolsep}{4pt}
  \begin{tabular}{@{}lcccccc@{}}
    \toprule
     & \multicolumn{2}{c}{ConvNeXtV2-Nano} & \multicolumn{2}{c}{DeiT-Tiny} & \multicolumn{2}{c}{Swin-Tiny} \\
    \cmidrule(lr){2-3} \cmidrule(lr){4-5} \cmidrule(lr){6-7}
    Method & Acc. $\uparrow$ & ECE $\downarrow$ & Acc. $\uparrow$ & ECE $\downarrow$ & Acc. $\uparrow$ & ECE $\downarrow$ \\
    \midrule
    Mixup+CutMix & $79.79 \pm 0.11$ & $9.48 \pm 0.13$ & $68.86 \pm 0.28$ & $12.34 \pm 0.27$ & $79.62 \pm 0.06$ & $7.50 \pm 0.13$ \\
    RICAP & $79.08 \pm 0.10$ & $5.95 \pm 0.17$ & $71.71 \pm 0.04$ & $2.04 \pm 0.28$ & $78.68 \pm 0.07$ & $7.78 \pm 0.05$ \\
    TreemapMix-SCE & $80.01 \pm 0.09$ & $\mathbf{2.74 \pm 0.27}$ & $71.78 \pm 0.26$ & $\mathbf{1.32 \pm 0.26}$ & $80.55 \pm 0.09$ & $\mathbf{2.52 \pm 0.11}$ \\
    TreemapMix-PL & $\mathbf{80.45 \pm 0.10}$ & $6.82 \pm 0.07$ & $\mathbf{71.81 \pm 0.07}$ & $2.71 \pm 0.15$ & $\mathbf{80.57 \pm 0.12}$ & $6.58 \pm 0.06$ \\
    \bottomrule
  \end{tabular}
\end{table*}

ImageNet-LT~\citep{liu2019largescale} is a long-tailed subset of ImageNet-1K, constructed by subsampling its training images to impose a power-law class-frequency distribution across the same 1,000 classes (5--1,280 images per class). For TreemapMix, the loader shuffles each class's images and forms rows by taking the next image from every class that still has one. It shuffles these rows and partitions each into $K$-source groups, preserving distinct labels. Incomplete groups are padded from earlier images in the same row, retaining only the required number of cyclic outputs; rows shorter than $K$ remain unmixed. Every image occurs once in the primary schedule, preserving the original class counts without minority-class oversampling; padding reuses images while keeping the total output count unchanged. Complete groups retain cyclic source-to-slot balance. Under the matched 110-epoch protocol, using three seeds, PL improves accuracy over Mixup+CutMix by 1.40, 1.02, 1.31, and 1.82 points; SCE's calibration advantage does not persist (Table~\ref{tab:long-tail-results}).

\begin{table*}[!htbp]
  \caption{ImageNet-LT accuracy and ECE under the matched 110-epoch protocol. Mean $\pm$ sample standard deviation over three seeds; ECE uses 15 bins. Bold marks the best mean per metric.}
  \label{tab:long-tail-results}
  \centering
  \footnotesize
  \setlength{\tabcolsep}{4pt}
  \begin{tabular}{@{}lcccc@{}}
    \toprule
    Method & ViT-Wee & ViT-Little & ViT-Medium & ViT-Betwixt \\
    \midrule
    \multicolumn{5}{c}{Top-1 accuracy (\%) $\uparrow$} \\
    \midrule
    Mixup+CutMix & $41.72 \pm 0.08$ & $41.67 \pm 0.43$ & $40.95 \pm 0.07$ & $39.78 \pm 0.21$ \\
    RICAP & $35.70 \pm 0.15$ & $35.39 \pm 0.00$ & $34.35 \pm 0.02$ & $31.44 \pm 3.37$ \\
    TreemapMix-SCE & $41.44 \pm 0.40$ & $41.09 \pm 0.40$ & $39.48 \pm 0.13$ & $38.67 \pm 0.27$ \\
    TreemapMix-PL & $\mathbf{43.12 \pm 0.06}$ & $\mathbf{42.69 \pm 0.07}$ & $\mathbf{42.26 \pm 0.07}$ & $\mathbf{41.60 \pm 0.17}$ \\
    \midrule
    \multicolumn{5}{c}{ECE (percentage points) $\downarrow$} \\
    \midrule
    Mixup+CutMix & $\mathbf{0.58 \pm 0.14}$ & $\mathbf{0.83 \pm 0.09}$ & $\mathbf{2.14 \pm 0.61}$ & $\mathbf{3.31 \pm 0.96}$ \\
    RICAP & $28.50 \pm 0.26$ & $31.86 \pm 0.21$ & $33.12 \pm 0.32$ & $30.89 \pm 2.81$ \\
    TreemapMix-SCE & $10.65 \pm 0.04$ & $13.00 \pm 0.56$ & $11.85 \pm 0.33$ & $11.35 \pm 0.62$ \\
    TreemapMix-PL & $17.90 \pm 0.63$ & $25.63 \pm 0.38$ & $23.62 \pm 0.74$ & $25.45 \pm 0.79$ \\
    \bottomrule
  \end{tabular}
\end{table*}

\paragraph{Computational performance.}\label{sec:runtime}
Table~\ref{tab:augmentation-loader-runtime} reports throughput, total loader time, and peak host memory from the review-time benchmark. Relative to single-image augmentation, SCE ($K=4$) reduces throughput by 19.0\% and PL ($K=6$) by 25.6\%, with increased host memory. These CPU-side measurements exclude model computation and do not establish end-to-end training overhead; prefetching can overlap preparation with GPU work (Appendix~\ref{app:runtime-overhead}).

\begin{table*}[!htbp]
  \centering
  \footnotesize
  \setlength{\tabcolsep}{4pt}
  \caption{Complete loader benchmark: five runs (seeds 42--46), batch size 128, eight workers, first batch, 20 warmup and 100 measured batches. Reduction is $100(1-T/T_{\mathrm{single}})$; $\Delta$RSS uses unrounded means. Memory is GiB ($2^{30}$ bytes). Total time covers all consumed batches; throughput uses measured batches only.}
  \label{tab:augmentation-loader-runtime}
  \begin{tabular}{@{}lcccc@{}}
    \toprule
    Method & $K$ & \makecell{Throughput\\(images/s)} & \makecell{Throughput\\reduction} & \makecell{Total loader\\time (s)} \\
    \midrule
    No Aug & 1 & $1670.4 \pm 30.2$ & $-30.2\%$ & $11.43 \pm 0.13$ \\
    Single-Image Aug & 1 & $1282.8 \pm 35.1$ & $0.0\%$ & $14.35 \pm 0.27$ \\
    Mixup+CutMix & 2 & $1223.4 \pm 29.7$ & $4.6\%$ & $15.13 \pm 0.36$ \\
    TreemapMix-SCE & 4 & $1039.4 \pm 37.0$ & $19.0\%$ & $19.83 \pm 0.46$ \\
    TreemapMix-PL & 6 & $954.3 \pm 33.0$ & $25.6\%$ & $21.30 \pm 0.36$ \\
    RICAP & 4 & $364.8 \pm 5.9$ & $71.6\%$ & $44.00 \pm 0.47$ \\
    \bottomrule
  \end{tabular}
  \par\medskip
  \begin{tabular}{@{}lccc@{}}
    \toprule
    Method & Peak RSS (GiB) & $\Delta$RSS (GiB) & Peak USS (GiB) \\
    \midrule
    No Aug & $28.49 \pm 0.07$ & $+0.11$ & $17.82 \pm 0.06$ \\
    Single-Image Aug & $28.38 \pm 0.16$ & $+0.00$ & $17.76 \pm 0.17$ \\
    Mixup+CutMix & $28.46 \pm 0.12$ & $+0.08$ & $17.82 \pm 0.15$ \\
    TreemapMix-SCE & $30.85 \pm 0.34$ & $+2.47$ & $20.16 \pm 0.28$ \\
    TreemapMix-PL & $32.41 \pm 0.52$ & $+4.04$ & $21.74 \pm 0.49$ \\
    RICAP & $36.82 \pm 0.14$ & $+8.44$ & $25.18 \pm 0.13$ \\
    \bottomrule
  \end{tabular}
\end{table*}

\paragraph{Sensitivity and longer training.}\label{sec:long-horizon}

The 320-epoch Wee/Little comparison (20 warmup epochs followed by 300 regular epochs) retains PL's highest mean accuracy and SCE's lowest ECE, with smaller PL gains over Mixup+CutMix of 0.91/0.21 points (Table~\ref{tab:in1k-long-horizon-results}). It reuses short-schedule settings. Single-seed optimizer and $(K,\alpha)$ studies appear in Appendices~\ref{app:optimizer-robustness} and \ref{app:in1k-hparam-sensitivity}. The latter supports small-to-medium $K$ and small $\alpha$ in the recommended regimes, without isolating filtering or scheduling effects.

\begin{table*}[t]
  \caption{ImageNet-1K long-horizon results for ViT-Wee and ViT-Little under a 320-epoch schedule. Accuracy is reported in percent, and ECE in percentage points using 15 bins. Values are mean $\pm$ one sample standard deviation over seeds 42, 43, and 44. Bold marks the best mean for each model and metric.}
  \label{tab:in1k-long-horizon-results}
  \centering
  \footnotesize
  \setlength{\tabcolsep}{4pt}
  \begin{tabular}{@{}lcccc@{}}
    \toprule
     & \multicolumn{2}{c}{ViT-Wee} & \multicolumn{2}{c}{ViT-Little} \\
    \cmidrule(lr){2-3} \cmidrule(lr){4-5}
    Method & Acc. $\uparrow$ & ECE $\downarrow$ & Acc. $\uparrow$ & ECE $\downarrow$ \\
    \midrule
    Single-Image Aug & $80.37 \pm 0.08$ & $3.89 \pm 0.16$ & $81.12 \pm 0.04$ & $3.39 \pm 0.06$ \\
    Mixup+CutMix & $79.73 \pm 0.15$ & $6.87 \pm 0.64$ & $81.74 \pm 0.09$ & $5.47 \pm 0.52$ \\
    TreemapMix-SCE & $80.41 \pm 0.04$ & $\mathbf{3.23 \pm 0.49}$ & $81.58 \pm 0.33$ & $\mathbf{2.77 \pm 1.42}$ \\
    TreemapMix-PL & $\mathbf{80.64 \pm 0.16}$ & $6.94 \pm 0.14$ & $\mathbf{81.95 \pm 0.34}$ & $7.44 \pm 0.54$ \\
    \bottomrule
  \end{tabular}
\end{table*}

\section{Diagnostic and Transfer Analysis}
We examine whether logits reflect the known region-area ordering and whether pretrained features transfer to COCO detection and instance segmentation. Hyperparameter sweeps and qualitative attributions appear in Appendices~\ref{app:in1k-hparam-sensitivity} and \ref{app:qualitative-visualizations}.

\begin{figure*}[t]
\centering
\includegraphics{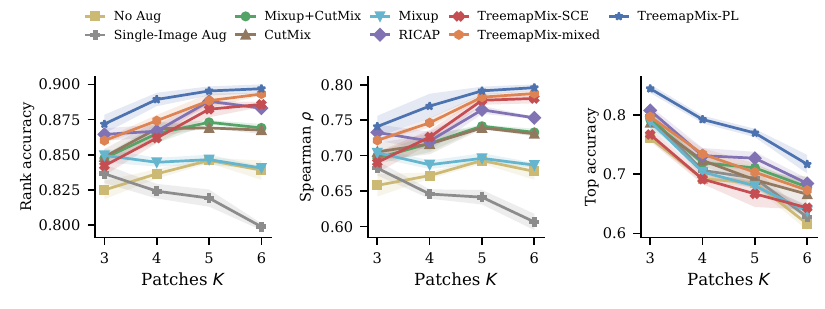}
\caption{Area-logit alignment diagnostics on the held-out composed diagnostic set for ViT-Betwixt checkpoints from the main ImageNet-1K study. Lines show means over three seeds and shaded bands denote $\pm 1$ sample standard deviation: pairwise rank accuracy (left), area-logit Spearman correlation (center), and largest-area top-class accuracy (right).}
\label{fig:area-logit-diagnostic}
\end{figure*}

\paragraph{Compositional Logit Alignment:}\label{sec:logit-alignment}
We evaluate the ViT-Betwixt checkpoints from the main ImageNet-1K comparison across three seeds on 500 held-out TreemapMix mixtures for each $K\in\{3,4,5,6\}$. Figure~\ref{fig:area-logit-diagnostic} reports three diagnostics restricted to classes present in each mixture: pairwise rank accuracy is the fraction of class pairs whose logits follow their area ordering, counting only pairs whose area fractions differ by at least 0.05 (left); Spearman correlation measures agreement between area and logit ranks, excluding samples with tied areas or logits (center); and largest-area top-class accuracy is the fraction of samples whose largest-area class has the highest logit among present classes, excluding tied largest areas (right).

TreemapMix-PL gives the highest pairwise rank accuracy, area-logit Spearman
correlation, and largest-area top-class accuracy in
Figure~\ref{fig:area-logit-diagnostic}. RICAP is competitive across all three
metrics and outperforms TreemapMix-SCE on largest-area top-class accuracy
at every evaluated $K$, despite also using area-weighted soft targets.
These results suggest that the ranking objective contributes to area-logit
alignment beyond the effect of compositional augmentation alone.

\paragraph{Feature Transfer to Instance Segmentation:}

We fine-tune ViT-Wee and ViT-Betwixt ImageNet-1K checkpoints with a ViTDet-style adapter and Mask R-CNN~\citep{liExploringPlainVision2022,heMaskRCNN2018} on COCO for 30 epochs at $256\times256$ resolution. Mixup+CutMix and RICAP provide the comparison baselines. Table~\ref{tab:coco-transfer-results} reports box/mask AP averaged over three transferred seeds; Appendix~\ref{app:det-hparams} provides fine-tuning hyperparameters and additional AP50 results (Table~\ref{tab:coco-transfer-ap50}).

\begin{table*}[t]
  \centering
  \small
\caption{COCO transfer results after ViTDet-style Mask R-CNN fine-tuning from
the main ImageNet-1K experiment checkpoints with three seeds per method.
Values are mean $\pm$ standard deviation over these three transferred seeds;
higher is better. Box AP evaluates bounding boxes and mask AP evaluates instance masks. AP is averaged over IoU thresholds from 0.50 to 0.95 in
increments of 0.05.}
  \label{tab:coco-transfer-results}
  \begin{tabular}{lcccc}
    \toprule
    & \multicolumn{2}{c}{ViT-Wee} & \multicolumn{2}{c}{ViT-Betwixt} \\
    \cmidrule(lr){2-3} \cmidrule(lr){4-5}
    Method & Box AP $\uparrow$ & Mask AP $\uparrow$ & Box AP $\uparrow$ & Mask AP $\uparrow$ \\
    \midrule
    \makecell[l]{Mixup+CutMix} & \makecell{23.66 $\pm$ 0.01} & \makecell{20.34 $\pm$ 0.10} & \makecell{29.04 $\pm$ 0.07} & \makecell{25.25 $\pm$ 0.10} \\
    RICAP & \makecell{23.08 $\pm$ 0.17} & \makecell{19.82 $\pm$ 0.07} & \makecell{27.86 $\pm$ 0.08} & \makecell{24.21 $\pm$ 0.06} \\
    \makecell[l]{TreemapMix-SCE} & \makecell{24.00 $\pm$ 0.17} & \makecell{20.55 $\pm$ 0.20} & \makecell{29.44 $\pm$ 0.16} & \makecell{25.56 $\pm$ 0.14} \\
    \makecell[l]{TreemapMix-PL} & \makecell{\bfseries 24.32 $\pm$ 0.12} & \makecell{\bfseries 20.71 $\pm$ 0.04} & \makecell{\bfseries 29.71 $\pm$ 0.09} & \makecell{\bfseries 25.75 $\pm$ 0.10} \\
    \bottomrule
  \end{tabular}
\end{table*}

Both TreemapMix variants improve over Mixup+CutMix and RICAP on box AP and
mask AP for both backbones, with TreemapMix-PL achieving the highest scores.
These results show that TreemapMix-pretrained checkpoints transfer favorably to
COCO detection and instance segmentation under this fine-tuning protocol.
However, they should not be interpreted as proving that the area-logit
alignment mechanism causes the downstream gains, since the transfer experiment
compares pretrained checkpoints rather than isolating individual representation
properties.

\section{Conclusion and Limitations}\label{sec:limitations}
We introduced TreemapMix, a Dirichlet-controlled $K$-way augmentation with area-proportional treemap layouts. Cyclic views preserve sample count and, for complete groups, balance each source's exposure to region sizes and ranks. The same composition supports area-weighted soft cross-entropy for probability matching and Plackett--Luce ranking for ordinal supervision. Under matched ImageNet-1K training recipes, SCE gives the strongest calibration, while PL gives the strongest Top-1 accuracy and area-logit alignment. Experiments extend this evaluation to CIFAR-100, additional architectures, long-tailed recognition, 320-epoch training, and loader efficiency. COCO transfer further demonstrates useful detection and instance-segmentation representations. Our comparisons leave the best attainable TreemapMix performance under separately optimized hyperparameters unexplored. Future work can tune recipes for longer schedules, study semantic content alongside region area, and isolate the contributions of cyclic views and class-balanced grouping. Other directions include calibration under skewed class priors, end-to-end training efficiency, adaptive grouping and schedules for $K$ and $\alpha$, and extensions to contrastive or self-supervised learning.

\begin{ack}
This work is supported by the New Generation Artificial Intelligence–National Science and Technology Major Project (No. 2026ZD0124304) and the National Natural Science Foundation of China (NSFC) under Grant No. 92582109. We also thank Tencent for providing GPU computing resources.

The authors declare no competing interests.
\end{ack}

\clearpage
\bibliographystyle{plainnat}
\bibliography{references}

\clearpage
\appendix
\section{TreemapMix Augmentation Details}\label{app:labelmix-details}

This appendix provides implementation details omitted from the main Method section. We use the notation introduced in Section~\ref{sec:method}.

\begin{algorithm}[!htbp]
\caption{TreemapMix augmentation}
\label{alg:labelmix}
\small
\begin{algorithmic}[1]
\Require dataset \(\mathcal{D}\), group size \(K\), concentration \(\alpha>0\), image size \(H\times W\), with \(1\le K\le\min(C,W)\) for this balanced-row protocol
\State build class-balanced rows and split them into groups of size \(K\)
\For{each group \(\mathcal{G}=\{(x_k,y_k)\}_{k=1}^{K}\)}
    \State sample \(w\sim\operatorname{Dirichlet}(\alpha,\ldots,\alpha)\)
    \State sort weights into \(\lambda_1\le\cdots\le\lambda_K\)
    \State construct boxes \(B_1,\dots,B_K\) with \(|B_i|/(HW)\approx\lambda_i\)
    \State apply a random valid symmetry transform to the boxes
    \For{\(s=0,\ldots,K-1\)}
        \State assign each source image \(x_{\pi_s(i)}\) to slot \(i\), for \(i=1,\dots,K\)
        \State render \(\tilde{x}^{(s)}=\sum_i M_i\odot\operatorname{Resize}(x_{\pi_s(i)},B_i)\)
        \State output \(\bigl(\tilde{x}^{(s)},\{(y_{\pi_s(i)},\lambda_i)\}_{i=1}^{K}\bigr)\)
    \EndFor
\EndFor
\end{algorithmic}
\end{algorithm}

\subsection{Class-balanced row construction}

Let the training dataset be \(\mathcal{D}=\{(x_n,y_n)\}_{n=1}^{N_{\mathrm{data}}}\), where \(y_n\in\{1,\dots,C\}\), and let \(\mathcal{I}_c=\{n:y_n=c\}\) be the set of dataset indices belonging to class $c$. A class-balanced row is constructed by sampling one example from each class. For imbalanced datasets, minority classes may be sampled with replacement, while majority classes may be subsampled across rows, so that every class contributes equally to each row.

Each row is shuffled and split into groups of size $K$. When $K\le C$, complete groups formed from a class-balanced row contain distinct labels by construction, because each class appears at most once in the row. This encourages diverse multi-class mixtures.

\subsection{Incomplete groups and row-size preservation}

Suppose a row contains $N_{\mathrm{row}}$ samples. Write \(N_{\mathrm{row}}=K\left\lfloor\frac{N_{\mathrm{row}}}{K}\right\rfloor+r\), where \(0\le r<K\). TreemapMix processes the first \(K\left\lfloor\frac{N_{\mathrm{row}}}{K}\right\rfloor\)
samples as complete groups of size $K$. If $r>0$, we select $K-r$ distinct image indices uniformly from the earlier, already processed part of the same row and load those images again to pad the tail group. We sample images, not new classes; because a balanced row contains one image per class, these padding images have labels distinct from the tail and from one another. Rows shorter than $K$ are passed through unmixed.

TreemapMix generates $K$ cyclic views from the temporary group but retains only the first $r$ outputs. Therefore, the total number of final TreemapMix samples contributed by the row is \(K\left\lfloor\frac{N_{\mathrm{row}}}{K}\right\rfloor+r=N_{\mathrm{row}}\).

Thus, output count matches the constructed row stream. Retaining only $r<K$ views does not provide the full-orbit balancing guarantee.

\subsection{Integer box conversion and layout fallback}

The squarified treemap is first computed in normalized coordinates. Let \(\bar{B}_i=(\bar{x}_i,\bar{y}_i,\bar{w}_i,\bar{h}_i)\) denote the normalized rectangle for slot $i$. We convert it to an integer pixel box \(B_i=(x_i^{(0)},y_i^{(0)},x_i^{(1)},y_i^{(1)})\). Because independent rounding can introduce gaps or overlaps, shared boundaries are snapped consistently. Including left/top boundaries and excluding right/bottom boundaries, a validated conversion satisfies \(B_i\cap B_j=\emptyset\) for \(i\neq j\) and \(\bigcup_{i=1}^{K}B_i=[0,W)\times[0,H)\).

A converted layout is considered valid if every box has positive area. Equivalently, for integer boxes, every box must have width and height at least one pixel. If the treemap layout fails this validity check, TreemapMix falls back to a stripe layout. The stripe layout partitions the canvas along one axis according to the sorted weights $\lambda_1,\dots,\lambda_K$, then applies the same integer-boundary snapping procedure. For vertical stripes, $K\le W$ and $H\ge1$ permit a positive-area full-canvas tiling. Boundaries are adjusted to give each stripe at least one pixel. Consequently, actual area fractions only approximate the sampled weights; minimum-width adjustments can exceed nearest-integer rounding error. Target weights remain the sampled weights.

\subsection{Symmetry transformations}

For square images, TreemapMix samples from the dihedral group
\[
\begin{aligned}
D_4=\{&
\text{identity},\space
90^\circ\text{ rotation},\space
180^\circ\text{ rotation},\space
270^\circ\text{ rotation},\\
&
\text{horizontal flip},\space
\text{vertical flip},\space
\text{transpose},\space
\text{anti-transpose}
\}.
\end{aligned}
\]
For non-square images, TreemapMix uses the shape-preserving subset
\[
\{
\text{identity},\space
180^\circ\text{ rotation},\space
\text{horizontal flip},\space
\text{vertical flip}
\}.
\]
These transformations preserve region areas, so they increase spatial diversity without changing the TreemapMix target.

\subsection{Cyclic views and slot exposure}\label{app:cyclic-views}

Cyclic views are a tractable alternative to enumerating all $K!$ image-to-slot assignments. If an epoch contains $G$ complete TreemapMix groups, full permutation enumeration would produce $K!G$ mixed images, increasing the sample count by \(\frac{K!G}{KG}=(K-1)!\). In contrast, cyclic TreemapMix produces exactly $K$ views per group, giving $KG$ mixed images and preserving the number of training samples after class-balanced row construction.

Cyclic permutation is designed to decouple source labels from fixed spatial slots. Although any individual mixed view may assign a class to a small or large region, the full cyclic orbit exposes every source image to every slot exactly once. Therefore, when the source labels in a group are distinct, every participating class is paired once with every area weight and, for PL training, every rank position induced by the sorted slot weights. Appendix Lemma~\ref{lem:cyclic-balancing} formalizes the resulting slot-exposure balance.

\newpage
\section{Soft TreemapMix Cross-Entropy Derivations}\label{app:soft-ce}

We use the sparse target notation from Section~\ref{sec:soft-ce}. For one TreemapMix sample, let \(\mathcal{T}=\{(c_i,\eta_i)\}_{i=1}^{K}\), where \(\eta_i\ge 0\) and \(\sum_{i=1}^{K}\eta_i=1\),
where $c_i$ is the class assigned to slot $i$ and $\eta_i$ is the corresponding slot weight. For cyclic view $s$, we have $c_i=y_{\pi_s(i)}$ and $\eta_i=\lambda_i$.

\subsection{KL derivation}

The sparse target distribution is \(q_c=\sum_{i=1}^{K}\eta_i\mathbf{1}[c_i=c]\).
It is a valid probability distribution because
\[
\sum_{c=1}^{C}q_c
=
\sum_{c=1}^{C}
\sum_{i=1}^{K}
\eta_i\mathbf{1}[c_i=c]
=
\sum_{i=1}^{K}
\eta_i
\sum_{c=1}^{C}
\mathbf{1}[c_i=c]
=
\sum_{i=1}^{K}\eta_i
=
1.
\]
Hence $q=(q_1,\dots,q_C)$ lies on the probability simplex.

The KL divergence from $q$ to $p_\theta$ is
\[
D_{\mathrm{KL}}(q\|p_\theta)
=
\sum_{c=1}^{C}
q_c\log\frac{q_c}{p_\theta(c\mid\tilde{x})}
=
\sum_{c=1}^{C}q_c\log q_c
-
\sum_{c=1}^{C}q_c\log p_\theta(c\mid\tilde{x}).
\]
The first term is independent of $\theta$. Thus, minimizing the KL divergence is equivalent to minimizing soft cross-entropy. Substituting the sparse target gives
\[
\begin{aligned}
\mathcal{L}_{\mathrm{TM\text{-}CE}}
&=
-\sum_{c=1}^{C}
\left(
\sum_{i=1}^{K}
\eta_i\mathbf{1}[c_i=c]
\right)
\log p_\theta(c\mid\tilde{x}) \\
&=
-\sum_{i=1}^{K}
\eta_i\log p_\theta(c_i\mid\tilde{x}),
\end{aligned}
\]
which is the sparse form in \eqref{eq:lm-ce}. For a mini-batch of size $B$,
\[
\mathcal{L}_{\mathrm{TM\text{-}CE}}
=
-\frac{1}{B}
\sum_{b=1}^{B}
\sum_{i=1}^{K}
\eta_{b,i}\log p_\theta(c_{b,i}\mid\tilde{x}_b).
\]

\subsection{Gradient derivation}

For one TreemapMix sample, write
\[
\mathcal{L}_{\mathrm{TM\text{-}CE}}
=
-\sum_{c=1}^{C}q_c\log p_c,
\qquad
p_c=
\frac{\exp(z_c)}{\sum_{j=1}^{C}\exp(z_j)}.
\]
Since $\sum_{c=1}^{C}q_c=1$, the log-softmax identity gives
\[
\begin{aligned}
\log p_c &= z_c-\log\sum_{j=1}^{C}\exp(z_j),\\
\mathcal{L}_{\mathrm{TM\text{-}CE}}
&=-\sum_{c=1}^{C}q_c z_c+\log\sum_{j=1}^{C}\exp(z_j),\\
\frac{\partial\mathcal{L}_{\mathrm{TM\text{-}CE}}}{\partial z_i}
&=-q_i+\frac{\exp(z_i)}{\sum_{j=1}^{C}\exp(z_j)}=p_i-q_i.
\end{aligned}
\]
Substituting $q_i=\sum_j\eta_j\mathbf{1}[c_j=i]$ gives the sparse gradient. For a mini-batch of size $B$,
\[
\frac{\partial \mathcal{L}_{\mathrm{TM\text{-}CE}}}{\partial z_{b,c}}
=
\frac{1}{B}
\left[
p_\theta(c\mid\tilde{x}_b)
-
\sum_{i=1}^{K}
\eta_{b,i}\mathbf{1}[c_{b,i}=c]
\right].
\]

\newpage
\subsection{Random cyclic-view target variance}

\refstepcounter{lemma}\label{lem:ce-cyclic-variance}
\paragraph{Lemma~\thelemma\ (Random cyclic-view and cyclic-average target variance).}
Assume distinct source labels, an unfiltered draw $w\sim\operatorname{Dirichlet}(\alpha,\ldots,\alpha)$ with $\alpha>0$, and let \(q_c^{(s)}=\sum_{i=1}^{K}\lambda_i\mathbf{1}[y_{\pi_s(i)}=c]\) be the TreemapMix-SCE target mass for class $c$ in cyclic view $s$. Conditioned on the source labels, if $c$ is a participating class and $s$ is sampled uniformly from $\{0,\dots,K-1\}$ independently of $w$, then
\[
\mathbb{E}_{w,s}[q_c^{(s)}]=\frac{1}{K},
\qquad
\operatorname{Var}_{w,s}(q_c^{(s)})=\frac{K-1}{K^2(K\alpha+1)}.
\]
Moreover, averaging over the full cyclic orbit gives \(\frac{1}{K}\sum_{s=0}^{K-1}q_c^{(s)}=\frac{1}{K}\), so the cyclic-average target mass has zero variance conditioned on the source labels. For an absent class, \(q_c^{(s)}=0\) for all \(s\).

\begin{proof}

\textit{Random cyclic view.} Assume the $K$ source labels are distinct, and let $c=y_k$ be a participating class. For each cyclic view $s$, class $c$ appears in exactly one slot. Since the cyclic map visits every slot exactly once as $s$ ranges from $0$ to $K-1$, sampling $s$ uniformly makes the slot index containing class $c$ uniformly distributed over $\{1,\dots,K\}$. Therefore, \(q_c^{(s)}=\lambda_I\), where \(I\sim\operatorname{Uniform}\{1,\dots,K\}\).
Because $(\lambda_1,\dots,\lambda_K)$ is a sorted permutation of $(w_1,\dots,w_K)$, a uniformly selected sorted component $\lambda_I$ has the same marginal distribution as a uniformly selected unsorted component $w_J$. By exchangeability of the symmetric Dirichlet distribution, this marginal distribution is the same as that of any component $w_j$. Hence
\(\mathbb{E}_{w,s}[q_c^{(s)}]=\mathbb{E}[w_j]=\frac{1}{K}\).
For a symmetric Dirichlet distribution with concentration parameter $\alpha$ in each coordinate,
\(\operatorname{Var}(w_j)=\frac{K-1}{K^2(K\alpha+1)}\). Therefore, \(\operatorname{Var}_{w,s}(q_c^{(s)})=\operatorname{Var}(w_j)=\frac{K-1}{K^2(K\alpha+1)}\).

\textit{Full cyclic orbit.} The cyclic-average identity follows from Lemma~\ref{lem:cyclic-balancing}. When labels are distinct, \(m_c=1\) for every participating class, so \(\frac{1}{K}\sum_{s=0}^{K-1}q_c^{(s)}=\frac{1}{K}\).
Since this equality holds for every Dirichlet sample, the conditional variance of the cyclic-average target mass is zero. If $c$ is absent from the group, then no slot carries class $c$, and \(q_c^{(s)}=0\) for all \(s\).

\textit{Fixed-logit interpretation.} If logits are fixed, then \(\frac{\partial \mathcal{L}_{\mathrm{TM\text{-}CE}}}{\partial z_c}=p_c-q_c^{(s)}\)
has the same variance as \(q_c^{(s)}\), since $p_c$ is held fixed while $w$ and $s$ vary. This is a fixed-logit target-side calculation, not a formula for the stochastic network-gradient variance when logits depend on the mixed input. Layout rejection and uniform fallback alter the weight distribution, so the exact Dirichlet variance need not hold for the filtered training sampler. The full-orbit identity holds for any normalized weights reused across all views.

\end{proof}

\newpage
\section{TreemapMix Plackett--Luce Derivations}\label{app:pl}

The derivation assumes $K$ distinct target classes. Let $r_1\succ\cdots\succ r_K$ denote their descending order by area weight, set $L=K$, and write $\rho_{r_\ell}$ for the area weight assigned to class $r_\ell$. Let $\mathcal{U}=\{1,\dots,C\}\setminus\{r_1,\dots,r_K\}$ be the absent classes, with total predicted probability mass $p_{\mathcal U}=1-\sum_{j=1}^{K}p_{r_j}$.

\subsection{From the Plackett--Luce probability to TreemapMix-PL}

Let the model output logits \(z=f_\theta(\tilde{x})\in\mathbb{R}^{C}\) and softmax probabilities \(p_c=p_\theta(c\mid\tilde{x})\).
For the descending ranking $r_1\succ\cdots\succ r_L$, the top-$L$ Plackett--Luce probability with unranked residual mass is
\[
P_{\mathrm{PL}}(r_1\succ\cdots\succ r_L\mid\tilde{x})
=
\prod_{\ell=1}^{L}
\frac{
p_{r_\ell}
}{
p_{\mathcal U}
+
\sum_{j=\ell}^{L}p_{r_j}
}.
\]
The corresponding unweighted ListMLE-style negative log-likelihood is
\[
\mathcal{L}_{\mathrm{ListMLE}}
=
-\sum_{\ell=1}^{L}
\log
\frac{
p_{r_\ell}
}{
p_{\mathcal U}
+
\sum_{j=\ell}^{L}p_{r_j}
}.
\]
TreemapMix-PL weights each ranking term by its area weight $\rho_{r_\ell}$:
\[
\mathcal{L}_{\mathrm{TM\text{-}PL}}
=
-\sum_{\ell=1}^{L}
\rho_{r_\ell}
\log
\frac{
p_{r_\ell}
}{
p_{\mathcal U}
+
\sum_{j=\ell}^{L}p_{r_j}
}.
\]
This is the descending form in \eqref{eq:lm-pl-desc}. This weighted stage-loss surrogate is not generally a normalized likelihood over rankings. With equal weights it is a constant rescaling of the unweighted top-$K$ ListMLE loss. Equal-weight entries retain their source-slot order, including under uniform fallback.

\subsection{Ascending implementation form}

For implementation, we reverse the descending order. Define \(a_i=r_{L+1-i}\), \(\gamma_i=\rho_{a_i}\), and \(\gamma_1\le\gamma_2\le\cdots\le\gamma_L\).
Thus, $a_1$ is the weakest ranked class and $a_L$ is the strongest. Reindexing the descending objective gives
\[
\mathcal{L}_{\mathrm{TM\text{-}PL}}
=
-\sum_{i=1}^{L}
\gamma_i
\log
\frac{
p_{a_i}
}{
p_{\mathcal U}
+
\sum_{j=1}^{i}p_{a_j}
}.
\]
Using
\[
D_i
=
p_{\mathcal U}
+
\sum_{j=1}^{i}p_{a_j}
=
1-\sum_{j=i+1}^{L}p_{a_j},
\]
we obtain
\[
\mathcal{L}_{\mathrm{TM\text{-}PL}}
=
\sum_{i=1}^{L}
\gamma_i
\left[
\log D_i-\log p_{a_i}
\right],
\]
which is the ascending implementation form in \eqref{eq:lm-pl-asc}. For a mini-batch of size $B$, where sample $b$ has $L_b$ distinct ranked classes, the corresponding batch objective is
\[
\mathcal{L}_{\mathrm{TM\text{-}PL}}
=
\frac{1}{B}
\sum_{b=1}^{B}
\sum_{i=1}^{L_b}
\gamma_{b,i}
\left[
\log D_{b,i}
-
\log p_\theta(a_{b,i}\mid \tilde{x}_b)
\right].
\]

\newpage
\subsection{Gradient derivation}

We differentiate the ascending objective above, holding its target weights and rank assignments fixed.
Define the stronger-class probability mass \(R_i=\sum_{j=i+1}^{L}p_{a_j}\) and \(D_i=1-R_i\), and the stronger-class set \(T_i=\{a_{i+1},\dots,a_L\}\). First, \(\frac{\partial \log p_{a_i}}{\partial z_c}=\mathbf{1}[c=a_i]-p_c\).
Next,
\[
\begin{aligned}
\frac{\partial R_i}{\partial z_c}
&=
\sum_{j=i+1}^{L}
p_{a_j}
\left(
\mathbf{1}[c=a_j]-p_c
\right) \\
&=
p_c\mathbf{1}[c\in T_i]-p_cR_i \\
&=
p_c\left(\mathbf{1}[c\in T_i]-R_i\right).
\end{aligned}
\]
The denominator derivatives are therefore
\[
\begin{aligned}
\frac{\partial D_i}{\partial z_c}
&=-\frac{\partial R_i}{\partial z_c}
=p_c\left(R_i-\mathbf{1}[c\in T_i]\right),\\
\frac{\partial\log D_i}{\partial z_c}
&=\frac{p_c\left(R_i-\mathbf{1}[c\in T_i]\right)}{D_i}.
\end{aligned}
\]
For the $i$-th term $\ell_i=\gamma_i[\log D_i-\log p_{a_i}]$, we have
\[
\begin{aligned}
\frac{\partial \ell_i}{\partial z_c}
&=
\gamma_i
\left[
\frac{
p_c\left(R_i-\mathbf{1}[c\in T_i]\right)
}{
D_i
}
-
\mathbf{1}[c=a_i]
+
p_c
\right] \\
&=
\gamma_i
\left[
\frac{
p_c\left(R_i-\mathbf{1}[c\in T_i]+D_i\right)
}{
D_i
}
-
\mathbf{1}[c=a_i]
\right] \\
&=
\gamma_i
\left[
\frac{
p_c\left(1-\mathbf{1}[c\in T_i]\right)
}{
D_i
}
-
\mathbf{1}[c=a_i]
\right],
\end{aligned}
\]
because $R_i+D_i=1$. Summing over $i$ gives
\[
\frac{\partial \mathcal{L}_{\mathrm{TM\text{-}PL}}}{\partial z_c}
=
\sum_{i=1}^{L}
\gamma_i
\left[
\frac{
p_c\left(1-\mathbf{1}[c\in T_i]\right)
}{
D_i
}
-
\mathbf{1}[c=a_i]
\right],
\]
which is the gradient in \eqref{eq:lm-pl-grad}.

\subsection{Ordinal gradient structure}

The gradient above has a simple ordinal interpretation.

\subsubsection*{Unranked class}

Let $u\notin\{a_1,\dots,a_L\}$. Then \(\mathbf{1}[u\in T_i]=0\) and \(\mathbf{1}[u=a_i]=0\) for all $i$, so \(\frac{\partial \mathcal{L}_{\mathrm{TM\text{-}PL}}}{\partial z_u}=p_u\sum_{i=1}^{L}\frac{\gamma_i}{D_i}\ge 0\).
Thus, direct gradient descent on logits suppresses absent classes. Updates through shared network parameters need not move each individual logit in that direction.

\subsubsection*{Ranked class}

Let $c=a_m$ for some $m\in\{1,\dots,L\}$. For $i<m$, class $a_m$ belongs to the stronger set $T_i$, so those terms vanish. For $i=m$, the contribution is \(\gamma_m\left(\frac{p_{a_m}}{D_m}-1\right)\). For $i>m$, class $a_m$ is not in $T_i$ and is not equal to $a_i$, so the remaining contribution is \(p_{a_m}\sum_{i=m+1}^{L}\frac{\gamma_i}{D_i}\).
Therefore,
\[
\frac{\partial \mathcal{L}_{\mathrm{TM\text{-}PL}}}{\partial z_{a_m}}
=
\gamma_m
\left(
\frac{p_{a_m}}{D_m}-1
\right)
+
p_{a_m}
\sum_{i=m+1}^{L}
\frac{\gamma_i}{D_i}.
\]
For the strongest class $a_L$, the sum is empty and $D_L=1$, so \(\frac{\partial \mathcal{L}_{\mathrm{TM\text{-}PL}}}{\partial z_{a_L}}=\gamma_L(p_{a_L}-1)\le 0\).
Thus, direct logit gradient descent promotes the strongest ranked class unless $p_{a_L}=1$, suppresses unranked classes, and regulates intermediate ranked classes through competition with stronger classes.

\newpage
\subsection{Proof of the Dirichlet order-statistics result}

\refstepcounter{lemma}\label{lem:pl-order}
\paragraph{Lemma~\thelemma\ (Dirichlet order statistics and ranking emphasis).}
Assume distinct target labels and unfiltered symmetric Dirichlet weights with $\alpha>0$, so that $L=K$ and $\gamma_i=\lambda_i$. Let \(\mu_i(\alpha)=\mathbb{E}[\lambda_i]\) denote the expected $i$-th sorted Dirichlet weight. Holding logits and rank assignments fixed while varying the weights, the expected TreemapMix-PL gradients satisfy
\[
\begin{aligned}
\mathbb{E}_{w}\!\left[\frac{\partial\mathcal{L}_{\mathrm{TM\text{-}PL}}}{\partial z_{a_m}}\right]
&=\mu_m(\alpha)\left(\frac{p_{a_m}}{D_m}-1\right)
+p_{a_m}\sum_{i=m+1}^{K}\frac{\mu_i(\alpha)}{D_i},\\
\mathbb{E}_{w}\!\left[\frac{\partial\mathcal{L}_{\mathrm{TM\text{-}PL}}}{\partial z_u}\right]
&=p_u\sum_{i=1}^{K}\frac{\mu_i(\alpha)}{D_i},\qquad u\notin\{a_1,\dots,a_K\}.
\end{aligned}
\]

\begin{proof}

\textit{Expected gradient.} Assume the $K$ target labels are distinct, so that $L=K$ and the ascending PL weights satisfy \(\gamma_i=\lambda_i\). Let \(\mu_i(\alpha)=\mathbb{E}[\lambda_i]\)
denote the expected $i$-th sorted Dirichlet weight. Holding logits and rank assignments fixed while varying the weights, $D_i$ and $p_c$ are fixed, and the ranked-class gradient is linear in $\lambda_i$. Taking expectations term by term replaces each $\lambda_i$ by $\mu_i(\alpha)$, giving both identities in the lemma.

The expected fixed-logit gradient therefore depends on $\alpha$ through $\mu_i(\alpha)$. This linearity argument does not establish monotonicity of individual gaps between order statistics.

\textit{Large-concentration limit.} For $\alpha\to\infty$, each coordinate has mean $1/K$ and variance $(K-1)/(K^2(K\alpha+1))\to0$. A union bound implies $\max_j|w_j-1/K|\to0$ in probability, and sorting preserves this convergence. Thus the loss tends to $1/K$ times the unweighted top-$K$ objective for fixed logits and rank assignments.

\textit{Scope.} For filtered sampling the same linearity uses moments of the actual accepted/fallback distribution, rather than unfiltered Dirichlet moments. These calculations hold predictions fixed; they do not describe the full training-gradient expectation when predictions vary with the mixed input.

\end{proof}

\newpage
\section{Proof of Cyclic Permutation Balancing}\label{app:cyclic-balancing}

\refstepcounter{lemma}\label{lem:cyclic-balancing}
\paragraph{Lemma~\thelemma\ (Cyclic permutation balances slot and target exposure)}

Let the normalized slot weights satisfy $\sum_{i=1}^{K}\lambda_i=1$, and define
\begin{equation}\label{eq:cyclic-soft-target}
q_c^{(s)}=\sum_{i=1}^{K}\lambda_i\mathbf{1}[y_{\pi_s(i)}=c],
\qquad m_c=\sum_{k=1}^{K}\mathbf{1}[y_k=c].
\end{equation}
Then, for every class $c$ and slot $i$,
\begin{equation}\label{eq:slot-balance}
\frac{1}{K}\sum_{s=0}^{K-1}\mathbf{1}[y_{\pi_s(i)}=c]=\frac{m_c}{K},
\end{equation}
and consequently
\begin{equation}\label{eq:cyclic-balance-soft}
\frac{1}{K}\sum_{s=0}^{K-1}q_c^{(s)}=\frac{m_c}{K}.
\end{equation}

\begin{proof}

\textit{Slot exposure.} For cyclic view $s$, slot $i$ receives source image $x_{\pi_s(i)}$ and label $y_{\pi_s(i)}$, where \(\pi_s(i)=1+\bigl((i+s-1)\bmod K\bigr)\).
For any fixed slot $i$, as $s$ ranges from $0$ to $K-1$, the index $\pi_s(i)$ visits every source position exactly once. Therefore, for any class $c$,
\(\sum_{s=0}^{K-1}\mathbf{1}[y_{\pi_s(i)}=c]=\sum_{k=1}^{K}\mathbf{1}[y_k=c]=m_c\). Dividing by $K$ gives the slot-balancing identity \(\frac{1}{K}\sum_{s=0}^{K-1}\mathbf{1}[y_{\pi_s(i)}=c]=\frac{m_c}{K}\),
which proves \eqref{eq:slot-balance}.

\textit{Target exposure.} The soft target mass assigned to class $c$ in cyclic view $s$ is \(q_c^{(s)}=\sum_{i=1}^{K}\lambda_i\mathbf{1}[y_{\pi_s(i)}=c]\).
Averaging over cyclic views gives
\[
\begin{aligned}
\frac{1}{K}
\sum_{s=0}^{K-1}
q_c^{(s)}
&=
\frac{1}{K}
\sum_{s=0}^{K-1}
\sum_{i=1}^{K}
\lambda_i\mathbf{1}[y_{\pi_s(i)}=c] \\
&=
\frac{1}{K}
\sum_{i=1}^{K}
\lambda_i
\sum_{s=0}^{K-1}
\mathbf{1}[y_{\pi_s(i)}=c] \\
&=
\frac{1}{K}
\sum_{i=1}^{K}
\lambda_i m_c \\
&=
\frac{m_c}{K},
\end{aligned}
\]
because $\sum_{i=1}^{K}\lambda_i=1$. This proves \eqref{eq:cyclic-balance-soft}.

When the $K$ source labels are distinct, $m_c=1$ for every participating class and $m_c=0$ for every absent class. Therefore, each participating class appears exactly once in every slot across the $K$ cyclic views. Since slots are associated with sorted weights and rank positions, cyclic permutation also prevents any participating class from being permanently tied to a fixed TreemapMix weight or rank. The PL denominator terms remain prediction-dependent, so this balancing does not remove ordinal competition; it only removes persistent source-label-to-slot bias.

\end{proof}

\newpage
\section{Implementation Details}
\subsection{Experimental Protocol and Hyperparameters}
\label{app:appendix-aug-configs}

Tables~\ref{tab:timm-adaptations}, \ref{tab:baseline-hparams-aug}, and
\ref{tab:baseline-hparams-train} summarize the ImageNet-1K experimental
protocol and baseline hyperparameters. Table~\ref{tab:timm-adaptations}
summarizes the schedule-level adaptations applied to the reference
\texttt{timm} recipes for the main 110-epoch comparison.
Tables~\ref{tab:baseline-hparams-aug} and
\ref{tab:baseline-hparams-train} separate data-augmentation settings from
optimizer and training settings for readability.

The baseline recipes are adapted from established \texttt{timm} ImageNet-1K
training configurations~\citep{rw2019timm,262588213843476SearchingBetterVit}.
Because our main study uses a shorter 110-epoch budget than many reference
recipes, we adapt the training recipe: we disable
\texttt{aug\_repeats}, adjust the learning rate, disable gradient clipping, and keep the remaining regularization settings fixed across methods.
All methods therefore share the same architecture, data budget, optimizer schedule, and evaluation protocol. The comparison isolates augmentation and loss effects, but does not claim per-baseline optimality.

Runs were executed on H20 nodes with eight GPUs and 2 TB system RAM, with independent two-GPU jobs colocated by a custom scheduler. Across reported main and appendix experiments, training used approximately 5.5k GPU-hours, while peak observed host-memory use remained below 400 GB per node.

The main ImageNet-1K comparison uses a global batch size of 1024. We omit epochs and warmup epochs
from Tables~\ref{tab:baseline-hparams-aug} and
\ref{tab:baseline-hparams-train} because those values are held fixed across
runs. The augmentation table lists the ImageNet-1K augmentation recipes used
in the main comparisons.

\begin{table}[!htbp]
  \centering
  \setlength{\tabcolsep}{5pt}
  \caption{Adaptations to the reference \texttt{timm} ImageNet-1K recipes used
for the main 110-epoch comparison. We shorten the original long schedules,
disable augmentation repeats, switch to \texttt{bfloat16}, and retune the
learning rate; all other recipe settings are kept unchanged unless listed.}
  \label{tab:timm-adaptations}
  \begin{tabular}{llll}
    \toprule
    Model & Setting & Original & New \\
    \midrule
    \multicolumn{4}{l}{\textbf{ViT-Betwixt}} \\
    & \texttt{dtype} & \texttt{float16} & \texttt{bfloat16} \\
    & \texttt{aug\_repeats} & 3.0 & 0.0 \\
    & \texttt{warmup\_epochs} & 20 & 10 \\
    & \texttt{epochs} & 600 & 110 \\
    & \texttt{lr} & 1.2e-3 & 1.5e-3 \\
    \midrule
    \multicolumn{4}{l}{\textbf{ViT-Medium}} \\
    & \texttt{dtype} & \texttt{float16} & \texttt{bfloat16} \\
    & \texttt{aug\_repeats} & 3.0 & 0.0 \\
    & \texttt{clip\_grad} & 5.0 & \texttt{null} \\
    & \texttt{warmup\_epochs} & 20 & 10 \\
    & \texttt{epochs} & 600 & 110 \\
    & \texttt{lr} & 1.2e-3 & 1.5e-3 \\
    \midrule
    \multicolumn{4}{l}{\textbf{ViT-Little}} \\
    & \texttt{dtype} & \texttt{float16} & \texttt{bfloat16} \\
    & \texttt{aug\_repeats} & 3.0 & 0.0 \\
    & \texttt{clip\_grad} & 5.0 & \texttt{null} \\
    & \texttt{warmup\_epochs} & 20 & 10 \\
    & \texttt{epochs} & 600 & 110 \\
    & \texttt{lr} & 1.2e-3 & 1.5e-3 \\
    \midrule
    \multicolumn{4}{l}{\textbf{ViT-Wee}} \\
    & \texttt{dtype} & \texttt{float16} & \texttt{bfloat16} \\
    & \texttt{aug\_repeats} & 3.0 & 0.0 \\
    & \texttt{clip\_grad} & 5.0 & \texttt{null} \\
    & \texttt{warmup\_epochs} & 20 & 10 \\
    & \texttt{epochs} & 450 & 110 \\
    & \texttt{lr} & 1.2e-3 & 1.5e-3 \\
    \bottomrule
  \end{tabular}
\end{table}

\begin{table*}[!htbp]
  \centering
  \footnotesize
  \caption{ImageNet-1K augmentation settings. All recipes except No Aug include single-image transforms (cropping, flipping, color/AutoAugment-style transforms, and random erasing). $p$ is the application probability; $\alpha$ denotes the method's mixing parameter. Settings were checked against archived run configurations and the historical training implementation. OpenMixup options beyond $\alpha$ and $p$ are implementation defaults without recorded overrides. The archived No Aug label disables single-image transforms but retains Mixup+CutMix with equal selection probabilities. RICAP records the configured close parameter; its effective cutoff depends on the configured training horizon. All TreemapMix variants use aspect-ratio filtering with maximum ratio 20.}
  \label{tab:baseline-hparams-aug}
  \setlength{\tabcolsep}{4pt}
  \begin{tabular}{@{}lp{10cm}@{}}
    \toprule
    Method & Mixing settings \\
    \midrule
    No Aug & Single-image transforms off; Mixup+CutMix with $\alpha=0.8/1.0$, $p=1$ \\
    Single-Image Aug & No mixing \\
    Mixup & $\alpha=0.8$, $p=1$ \\
    CutMix & $\alpha=1.0$, $p=1$ \\
    Mixup+CutMix & $\alpha=0.8/1.0$, $p=1$; choose either method with probability $0.5$ \\
    RICAP & $p=0.5$; \texttt{close\_epochs}=10; center $[0.5,1.5]$; scale $[0.8,1.2]$ \\
    \midrule
    FMix & $\alpha=0.4$, $p=1$; spectral decay $3$, hard mask \\
    GridMix & $\alpha=0.1$, $p=1$; 20 holes; hole/cut aspect and cut-area ratios $1$ \\
    ResizeMix & $\alpha=0.1$ (unused); $p=1$; scale uniform in $[0.1,0.8]$, nearest interpolation \\
    SaliencyMix & $\alpha=0.25$, $p=1$ \\
    SmoothMix & $\alpha=0.6$, $p=1$ \\
    TokenMix & $\alpha=0.4$, $p=1$; block mask, minimum 14 tokens \\
    TLA & $\alpha=0.2$, $p=1$; patch size 16 \\
    \midrule
    TreemapMix-SCE & $K=4$; cosine $\alpha:0.1\to0.5$ \\
    TreemapMix-PL & $K=6$; cosine $\alpha:0.1\to0.5$ \\
    TreemapMix-mixed & $K=4$; cosine $\alpha:0.1\to0.5$; $\beta=0.5$ \\
    \bottomrule
  \end{tabular}
\end{table*}

\begin{table*}[!htbp]
  \centering
  \caption{ViT baseline optimizer and training hyperparameters for the main comparison. Opt. denotes optimizer, LR learning rate, WD
  weight decay, and EMA exponential moving average of model weights. NAdamW
  is Nesterov-accelerated Adam with decoupled weight decay. DropPath is the
  stochastic-depth probability.}
  \label{tab:baseline-hparams-train}
  \begin{tabular}{lcccccc}
    \toprule
    Model & Opt. & Base LR & LR Decay & WD & DropPath & EMA \\
    \midrule
    ViT-Wee & NAdamW & 1.5e-3 & cosine & 0.06 & 0.2 & yes \\
    ViT-Little & NAdamW & 1.5e-3 & cosine & 0.06 & 0.2 & no \\
    ViT-Medium & NAdamW & 1.5e-3 & cosine & 0.06 & 0.2 & no \\
    ViT-Betwixt & NAdamW & 1.5e-3 & cosine & 0.08 & 0.2 & no \\
    \bottomrule
  \end{tabular}%
\end{table*}

\subsection{Additional COCO transfer results}
\label{app:det-hparams}

This subsection supplements the main transfer analysis with the fine-tuning hyperparameters in Table~\ref{tab:det-hparams} and box/mask AP50 results in Table~\ref{tab:coco-transfer-ap50}. AP50 evaluates average precision at an intersection-over-union threshold of 0.50, complementing the main table's AP averaged over thresholds from 0.50 to 0.95.

\begin{table*}[!htbp]
  \centering
  \small
  \caption{Hyperparameters for the object-detection transfer experiments with
  a ViTDet-style adapter and Mask R-CNN on COCO. Embed Dim is the embedding
  dimension, Heads is the attention-head count, MLP Ratio is the multilayer
  perceptron expansion ratio, and LR is the learning rate.}
  \label{tab:det-hparams}
  \begin{tabular}{lccccccc}
    \toprule
    Model & Embed Dim & Depth & Heads & MLP Ratio & Drop Path & LR & Layer Decay \\
    \midrule
    ViT-Wee & 256 & 14 & 4 & 5.0 & 0.1 & 6.0e-4 & 0.7 \\
    ViT-Betwixt & 640 & 12 & 10 & 4.0 & 0.1 & 8.0e-4 & 0.7 \\
    \bottomrule
  \end{tabular}%
\end{table*}

At this fixed overlap threshold, PL improves box/mask AP50 over Mixup+CutMix by 1.23/0.95 points on ViT-Wee and 1.12/0.89 points on ViT-Betwixt (Table~\ref{tab:coco-transfer-ap50}). Thus, the transfer gains also appear under the less stringent AP50 criterion, complementing the main multi-threshold AP results.
\begin{table*}[!htbp]
  \centering
\caption{COCO transfer AP50 after ViTDet-style Mask R-CNN fine-tuning from
ImageNet-1K checkpoints. In this transfer setting, TreemapMix-PL obtains the
best box and mask AP50 for both ViT-Wee and ViT-Betwixt, while TreemapMix-SCE
also improves over Mixup+CutMix and RICAP. Values are mean $\pm$ standard
deviation over three seeds; higher is better, and bold marks the best mean in
each column.}
  \label{tab:coco-transfer-ap50}
  \begin{tabular}{lcccc}
    \toprule
    & \multicolumn{2}{c}{ViT-Wee} & \multicolumn{2}{c}{ViT-Betwixt} \\
    \cmidrule(lr){2-3} \cmidrule(lr){4-5}
    Method & Box AP50 $\uparrow$ & Mask AP50 $\uparrow$ & Box AP50 $\uparrow$ & Mask AP50 $\uparrow$ \\
    \midrule
    \makecell[l]{Mixup+CutMix} & 39.67 $\pm$ 0.01 & 35.64 $\pm$ 0.09 & 46.46 $\pm$ 0.19 & 42.59 $\pm$ 0.30 \\
    RICAP & 38.68 $\pm$ 0.24 & 34.82 $\pm$ 0.20 & 44.91 $\pm$ 0.04 & 41.05 $\pm$ 0.20 \\
    \makecell[l]{TreemapMix-SCE} & 40.37 $\pm$ 0.19 & 36.18 $\pm$ 0.19 & 47.06 $\pm$ 0.15 & 43.12 $\pm$ 0.13 \\
    \makecell[l]{TreemapMix-PL} & \textbf{40.90 $\pm$ 0.01} & \textbf{36.59 $\pm$ 0.13} & \textbf{47.58 $\pm$ 0.27} & \textbf{43.48 $\pm$ 0.14} \\
    \bottomrule
  \end{tabular}
\end{table*}
% Flush protocol and transfer tables before the self-contained loss-code page.
\clearpage
\subsection{Loss functions in code}
\label{app:loss-code}

We include concise PyTorch reference implementations of TreemapMix-SCE
(Figure~\ref{fig:treemapmix-sce-code}), TreemapMix-PL
(Figure~\ref{fig:treemapmix-pl-code}), and the mixed-loss variant
(Figure~\ref{fig:treemapmix-mixed-code}) for the distinct-label mathematical objectives. The PL listing is a numerically stable reference, not a reconstruction of every branch of the experimental training code.

% A single float keeps all three implementations and their captions together.
\begin{figure}[!htbp]
\centering
\input{fig/fig_treemapmix_sce_code}
\input{fig/fig_treemapmix_pl_code}
\input{fig/fig_treemapmix_mixed_code}
\end{figure}

\subsection{Runtime overhead}
\label{app:runtime-overhead}
For the configurations in this benchmark, augmentations are applied in the data loader.
Table~\ref{tab:augmentation-loader-runtime} isolates data-loader throughput
and host-memory usage for six augmentation configurations. The benchmark
measures CPU-side loading and augmentation without model forward or backward
passes. Peak resident set size (RSS) and unique set size (USS) are summed over
the loader process tree. RSS can count shared pages more than once, while USS
measures private memory. Memory is reported in GiB.

Relative to the single-image recipe (1282.8 images/s), TreemapMix-SCE at
$K=4$ reaches 1039.4 images/s, a 19.0\% throughput reduction;
TreemapMix-PL at $K=6$ reaches 954.3 images/s, a 25.6\% reduction.
The corresponding peak USS increases from 17.76 GiB to 20.16 and 21.74 GiB.
Mixup+CutMix reaches 1223.4 images/s, a 4.6\% reduction.
The RICAP configuration reaches 364.8 images/s with 25.18 GiB peak USS.
TreemapMix therefore adds measurable host-side cost.

The training configuration uses a prefetch factor of 2, which allows CPU
preparation to overlap GPU computation. The isolated loader benchmark does
not measure how much of this overhead is hidden during training, so it does
not establish an end-to-end per-epoch timing comparison.

\section{Additional Experiments}
\subsection{Full ImageNet-1K Results}
\label{app:in1k-confidence-results}

Tables~\ref{tab:in1k-full-results-acc} and
\ref{tab:in1k-full-results-ece} report the full ImageNet-1K
main-comparison Top-1 accuracy and ECE results in the main text, respectively. Figure~\ref{fig:in1k-main-results-top1} summarizes the same results graphically.

All 16 configurations are also evaluated using negative log-likelihood (NLL) and Brier score in Tables~\ref{tab:in1k-full-results-nll} and~\ref{tab:in1k-full-results-brier}. SCE has the lowest mean NLL on all four backbones; Brier differences are small and sometimes favor the mixed objective. These complementary metrics preserve the broader confidence-quality evaluation beyond ECE.

\begin{table*}[!htbp]
\centering
\footnotesize
\setlength{\tabcolsep}{4pt}
\caption{ImageNet-1K negative log-likelihood for all 16 configurations under the matched 110-epoch protocol. Values are mean $\pm$ one sample standard deviation over three seeds. Bold marks the lowest mean in each backbone.
}
\label{tab:in1k-full-results-nll}
\begin{tabular}{@{}lcccc@{}}
\toprule
Method & ViT-Wee $\downarrow$ & ViT-Little $\downarrow$ & ViT-Medium $\downarrow$ & ViT-Betwixt $\downarrow$ \\
\midrule
Single-Image Aug & $0.923 \pm 0.006$ & $0.869 \pm 0.008$ & $0.864 \pm 0.001$ & $0.855 \pm 0.001$ \\
No Aug & $1.120 \pm 0.009$ & $1.137 \pm 0.005$ & $1.154 \pm 0.006$ & $1.134 \pm 0.008$ \\
Mixup+CutMix & $1.015 \pm 0.011$ & $0.896 \pm 0.002$ & $0.854 \pm 0.008$ & $0.811 \pm 0.004$ \\
Mixup & $1.012 \pm 0.004$ & $0.904 \pm 0.002$ & $0.862 \pm 0.002$ & $0.820 \pm 0.002$ \\
CutMix & $0.994 \pm 0.011$ & $0.870 \pm 0.004$ & $0.840 \pm 0.005$ & $0.787 \pm 0.005$ \\
RICAP & $0.893 \pm 0.004$ & $0.846 \pm 0.007$ & $0.870 \pm 0.008$ & $0.857 \pm 0.003$ \\
FMix & $0.965 \pm 0.004$ & $0.850 \pm 0.001$ & $0.821 \pm 0.002$ & $0.793 \pm 0.001$ \\
GridMix & $0.994 \pm 0.005$ & $0.893 \pm 0.002$ & $0.865 \pm 0.006$ & $0.837 \pm 0.006$ \\
ResizeMix & $0.967 \pm 0.004$ & $0.851 \pm 0.007$ & $0.816 \pm 0.007$ & $0.784 \pm 0.006$ \\
SaliencyMix & $0.963 \pm 0.005$ & $0.855 \pm 0.003$ & $0.824 \pm 0.007$ & $0.791 \pm 0.002$ \\
SmoothMix & $1.008 \pm 0.009$ & $0.878 \pm 0.004$ & $0.840 \pm 0.002$ & $0.803 \pm 0.004$ \\
TokenMix & $0.993 \pm 0.011$ & $0.870 \pm 0.005$ & $0.829 \pm 0.004$ & $0.796 \pm 0.004$ \\
TLA & $0.956 \pm 0.010$ & $0.842 \pm 0.008$ & $0.806 \pm 0.003$ & $0.775 \pm 0.001$ \\
TreemapMix-mixed & $0.862 \pm 0.008$ & $0.778 \pm 0.011$ & $0.739 \pm 0.005$ & $0.728 \pm 0.003$ \\
TreemapMix-PL & $0.881 \pm 0.012$ & $0.803 \pm 0.001$ & $0.773 \pm 0.002$ & $0.766 \pm 0.008$ \\
TreemapMix-SCE & $\mathbf{0.857 \pm 0.003}$ & $\mathbf{0.775 \pm 0.002}$ & $\mathbf{0.735 \pm 0.004}$ & $\mathbf{0.714 \pm 0.007}$ \\
\bottomrule
\end{tabular}
\end{table*}

\begin{table*}[!htbp]
\centering
\footnotesize
\setlength{\tabcolsep}{4pt}
\caption{ImageNet-1K Brier score for all 16 configurations under the matched 110-epoch protocol. Values are mean $\pm$ one sample standard deviation over three seeds. Bold marks the lowest mean in each backbone.
}
\label{tab:in1k-full-results-brier}
\begin{tabular}{@{}lcccc@{}}
\toprule
Method & ViT-Wee $\downarrow$ & ViT-Little $\downarrow$ & ViT-Medium $\downarrow$ & ViT-Betwixt $\downarrow$ \\
\midrule
Single-Image Aug & $0.314 \pm 0.002$ & $0.294 \pm 0.003$ & $0.291 \pm 0.001$ & $0.285 \pm 0.001$ \\
No Aug & $0.355 \pm 0.003$ & $0.349 \pm 0.002$ & $0.344 \pm 0.001$ & $0.335 \pm 0.001$ \\
Mixup+CutMix & $0.342 \pm 0.003$ & $0.305 \pm 0.001$ & $0.292 \pm 0.003$ & $0.277 \pm 0.002$ \\
Mixup & $0.340 \pm 0.002$ & $0.308 \pm 0.001$ & $0.294 \pm 0.001$ & $0.280 \pm 0.001$ \\
CutMix & $0.338 \pm 0.003$ & $0.298 \pm 0.002$ & $0.287 \pm 0.002$ & $0.270 \pm 0.001$ \\
RICAP & $0.319 \pm 0.002$ & $0.303 \pm 0.002$ & $0.304 \pm 0.002$ & $0.299 \pm 0.001$ \\
FMix & $0.330 \pm 0.002$ & $0.293 \pm 0.000$ & $0.283 \pm 0.001$ & $0.273 \pm 0.001$ \\
GridMix & $0.335 \pm 0.002$ & $0.303 \pm 0.001$ & $0.294 \pm 0.002$ & $0.284 \pm 0.001$ \\
ResizeMix & $0.328 \pm 0.001$ & $0.293 \pm 0.002$ & $0.281 \pm 0.002$ & $0.269 \pm 0.001$ \\
SaliencyMix & $0.327 \pm 0.001$ & $0.293 \pm 0.001$ & $0.283 \pm 0.002$ & $0.272 \pm 0.000$ \\
SmoothMix & $0.340 \pm 0.003$ & $0.299 \pm 0.001$ & $0.288 \pm 0.001$ & $0.275 \pm 0.001$ \\
TokenMix & $0.336 \pm 0.003$ & $0.297 \pm 0.000$ & $0.285 \pm 0.001$ & $0.272 \pm 0.001$ \\
TLA & $0.327 \pm 0.003$ & $0.292 \pm 0.002$ & $0.280 \pm 0.001$ & $0.269 \pm 0.000$ \\
TreemapMix-mixed & $0.313 \pm 0.002$ & $\mathbf{0.288 \pm 0.003}$ & $\mathbf{0.275 \pm 0.001}$ & $0.269 \pm 0.000$ \\
TreemapMix-PL & $\mathbf{0.313 \pm 0.003}$ & $0.288 \pm 0.001$ & $0.277 \pm 0.001$ & $0.271 \pm 0.002$ \\
TreemapMix-SCE & $0.314 \pm 0.000$ & $0.289 \pm 0.001$ & $0.276 \pm 0.001$ & $\mathbf{0.268 \pm 0.002}$ \\
\bottomrule
\end{tabular}
\end{table*}

\subsection{CIFAR-100 Results}

\label{app:cifar100-results}
\begin{figure*}[!htbp]
  \centering
  \includegraphics{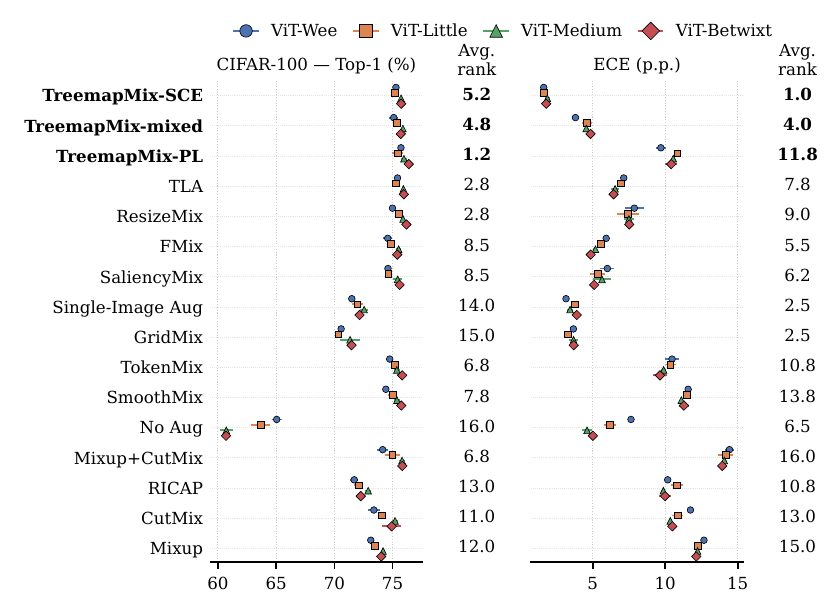}
  \caption{CIFAR-100 comparison at 256-pixel resolution under an approximately 210-epoch schedule, including 10 warmup epochs. Points show the mean over seeds 42, 43, and 44; intervals denote $\pm$ one sample standard deviation. ECE uses 15 bins and is reported in percentage points. Average ranks summarize within-backbone ranks, with rank 1 best. Complete values appear in Tables~\ref{tab:cifar100-full-results-acc} and \ref{tab:cifar100-full-results-ece}.}
  \label{fig:cifar100-results}
\end{figure*}

We evaluate all 16 augmentation settings on CIFAR-100 with the same four
ViT backbones and 256-pixel inputs. The matched schedule comprises 210 total epochs, including
10 warmup epochs. TreemapMix uses the settings from the primary comparison:
$K=4$ for SCE and the mixed objective, $K=6$ for PL, and a
cosine $\alpha$ schedule from $0.1$ to $0.5$.
Figure~\ref{fig:cifar100-results} summarizes Top-1 accuracy and ECE with 15 bins; Tables~\ref{tab:cifar100-full-results-acc} and \ref{tab:cifar100-full-results-ece} give all numerical results.

TreemapMix-SCE has the lowest ECE on all four backbones, ranging from
1.62 to 1.87 percentage points. TreemapMix-PL has the highest Top-1 accuracy
on ViT-Wee, Medium, and Betwixt, reaching 75.72\%, 75.97\%, and 76.39\%,
respectively. On ViT-Little, ResizeMix reaches 75.54\% compared with
75.42\% for PL. The mixed objective does not consistently improve on either
individual objective. These results retain the accuracy--calibration
trade-off, without a uniform accuracy advantage across every model.

\begin{table*}[!htbp]
  \centering
  \footnotesize
  \setlength{\tabcolsep}{4pt}
  \caption{CIFAR-100 Top-1 accuracy (\%) at 256 pixels, approximately 210 epochs including 10 warmup epochs. Mean $\pm$ sample standard deviation over three seeds; bold marks the best mean per backbone.}
  \label{tab:cifar100-full-results-acc}
  \begin{tabular}{@{}lcccc@{}}
    \toprule
    Method & ViT-Wee $\uparrow$ & ViT-Little $\uparrow$ & ViT-Medium $\uparrow$ & ViT-Betwixt $\uparrow$ \\
    \midrule
    No Aug & $65.05 \pm 0.44$ & $63.69 \pm 0.81$ & $60.72 \pm 0.54$ & $60.70 \pm 0.35$ \\
    Single-Image Aug & $71.50 \pm 0.09$ & $71.97 \pm 0.45$ & $72.56 \pm 0.35$ & $72.16 \pm 0.06$ \\
    Mixup+CutMix & $74.15 \pm 0.47$ & $74.96 \pm 0.64$ & $75.81 \pm 0.24$ & $75.83 \pm 0.16$ \\
    Mixup & $73.13 \pm 0.10$ & $73.46 \pm 0.34$ & $74.18 \pm 0.18$ & $74.02 \pm 0.18$ \\
    CutMix & $73.39 \pm 0.51$ & $74.05 \pm 0.16$ & $75.21 \pm 0.22$ & $74.92 \pm 0.83$ \\
    FMix & $74.58 \pm 0.37$ & $74.84 \pm 0.22$ & $75.52 \pm 0.22$ & $75.40 \pm 0.07$ \\
    GridMix & $70.58 \pm 0.16$ & $70.34 \pm 0.27$ & $71.34 \pm 0.84$ & $71.48 \pm 0.22$ \\
    ResizeMix & $74.99 \pm 0.22$ & $\mathbf{75.54 \pm 0.34}$ & $75.89 \pm 0.05$ & $76.18 \pm 0.12$ \\
    SaliencyMix & $74.60 \pm 0.14$ & $74.64 \pm 0.29$ & $75.43 \pm 0.40$ & $75.60 \pm 0.28$ \\
    SmoothMix & $74.42 \pm 0.11$ & $75.00 \pm 0.36$ & $75.38 \pm 0.23$ & $75.74 \pm 0.28$ \\
    TLA & $75.42 \pm 0.13$ & $75.27 \pm 0.13$ & $75.93 \pm 0.02$ & $75.96 \pm 0.18$ \\
    TokenMix & $74.75 \pm 0.28$ & $75.18 \pm 0.24$ & $75.36 \pm 0.25$ & $75.82 \pm 0.44$ \\
    RICAP & $71.71 \pm 0.35$ & $72.07 \pm 0.38$ & $72.90 \pm 0.27$ & $72.26 \pm 0.26$ \\
    TreemapMix-SCE & $75.29 \pm 0.11$ & $75.20 \pm 0.36$ & $75.72 \pm 0.11$ & $75.73 \pm 0.09$ \\
    TreemapMix-mixed & $75.10 \pm 0.37$ & $75.38 \pm 0.24$ & $75.89 \pm 0.15$ & $75.70 \pm 0.39$ \\
    TreemapMix-PL & $\mathbf{75.72 \pm 0.29}$ & $75.42 \pm 0.45$ & $\mathbf{75.97 \pm 0.29}$ & $\mathbf{76.39 \pm 0.35}$ \\
    \bottomrule
  \end{tabular}
\end{table*}

\begin{table*}[!htbp]
  \centering
  \footnotesize
  \setlength{\tabcolsep}{4pt}
  \caption{CIFAR-100 ECE (15 bins, percentage points). Same runs as Table~\ref{tab:cifar100-full-results-acc}; bold marks the lowest mean per backbone.}
  \label{tab:cifar100-full-results-ece}
  \begin{tabular}{@{}lcccc@{}}
    \toprule
    Method & ViT-Wee $\downarrow$ & ViT-Little $\downarrow$ & ViT-Medium $\downarrow$ & ViT-Betwixt $\downarrow$ \\
    \midrule
    No Aug & $7.64 \pm 0.20$ & $6.19 \pm 0.39$ & $4.61 \pm 0.32$ & $5.01 \pm 0.08$ \\
    Single-Image Aug & $3.16 \pm 0.17$ & $3.75 \pm 0.24$ & $3.44 \pm 0.22$ & $3.90 \pm 0.30$ \\
    Mixup+CutMix & $14.43 \pm 0.28$ & $14.16 \pm 0.50$ & $14.07 \pm 0.19$ & $13.93 \pm 0.22$ \\
    Mixup & $12.67 \pm 0.18$ & $12.24 \pm 0.24$ & $12.23 \pm 0.05$ & $12.14 \pm 0.24$ \\
    CutMix & $11.74 \pm 0.02$ & $10.85 \pm 0.37$ & $10.35 \pm 0.09$ & $10.49 \pm 0.28$ \\
    FMix & $5.93 \pm 0.14$ & $5.57 \pm 0.18$ & $5.19 \pm 0.14$ & $4.86 \pm 0.22$ \\
    GridMix & $3.67 \pm 0.21$ & $3.30 \pm 0.26$ & $3.68 \pm 0.33$ & $3.69 \pm 0.22$ \\
    ResizeMix & $7.87 \pm 0.66$ & $7.41 \pm 0.76$ & $7.50 \pm 0.34$ & $7.52 \pm 0.26$ \\
    SaliencyMix & $6.01 \pm 0.49$ & $5.35 \pm 0.52$ & $5.64 \pm 0.63$ & $5.10 \pm 0.16$ \\
    SmoothMix & $11.58 \pm 0.10$ & $11.50 \pm 0.25$ & $11.11 \pm 0.12$ & $11.29 \pm 0.33$ \\
    TLA & $7.14 \pm 0.10$ & $6.93 \pm 0.12$ & $6.54 \pm 0.29$ & $6.44 \pm 0.31$ \\
    TokenMix & $10.47 \pm 0.47$ & $10.35 \pm 0.37$ & $9.89 \pm 0.17$ & $9.64 \pm 0.45$ \\
    RICAP & $10.17 \pm 0.11$ & $10.81 \pm 0.44$ & $9.87 \pm 0.14$ & $9.99 \pm 0.42$ \\
    TreemapMix-SCE & $\mathbf{1.62 \pm 0.20}$ & $\mathbf{1.62 \pm 0.20}$ & $\mathbf{1.87 \pm 0.17}$ & $\mathbf{1.79 \pm 0.08}$ \\
    TreemapMix-mixed & $3.81 \pm 0.20$ & $4.60 \pm 0.35$ & $4.55 \pm 0.18$ & $4.85 \pm 0.27$ \\
    TreemapMix-PL & $9.70 \pm 0.32$ & $10.83 \pm 0.14$ & $10.55 \pm 0.11$ & $10.41 \pm 0.40$ \\
    \bottomrule
  \end{tabular}
\end{table*}

\subsection{Optimizer Sensitivity}
\label{app:optimizer-robustness}
We next test whether the gains of TreemapMix persist across different
optimizers on ImageNet-1K. We compare Muon
\citep{jordan2024muon}, AdamW (Adam with decoupled weight decay)
\citep{loshchilovDecoupledWeightDecay2019}, and NAdamW
\citep{Dahl2023AlgoPerf,Kasimbeg2025AlgoPerfResults}. We keep the core
TreemapMix hyperparameters fixed across optimizers. Because the main
experiments use NAdamW with a learning rate of $1.5\times10^{-3}$, we retain
this value for NAdamW and also use it for AdamW to keep the Adam-style
optimizer comparison matched. For Muon, we use a separately tuned learning
rate of $5\times10^{-3}$.

Table~\ref{tab:in1k-optimizer-results} compares Mixup+CutMix,
TreemapMix-SCE, and TreemapMix-PL under AdamW, Muon, and NAdamW for
the settings in the caption. These results should be interpreted as an
optimizer-sensitivity check rather than a definitive optimizer ranking. Across
the tested optimizers, the main qualitative pattern from the primary
ImageNet-1K comparison is preserved: TreemapMix-PL generally gives the highest
Top-1 accuracy, whereas TreemapMix-SCE gives the lowest ECE in most settings.
Muon gives the strongest Top-1 accuracy in several configurations, but this
gain is often accompanied by higher ECE. AdamW and NAdamW are more competitive
on calibration, although neither dominates across both metrics. Overall, these
single-seed results suggest that TreemapMix's accuracy and calibration trends
are not specific to one optimizer, while the absolute accuracy-calibration
trade-off still depends on the optimizer.

\begin{table*}[!htbp]
  \centering
\caption{ImageNet-1K optimizer comparison under the settings in Appendix~\ref{app:optimizer-robustness}, using one run with seed 42 per configuration. Accuracy is reported in percent and ECE in percentage
points; bold marks the best value for each metric within each row.}
  \label{tab:in1k-optimizer-results}
  \begin{tabular}{llcccccc}
    \toprule
    & & \multicolumn{2}{c}{Mixup+CutMix}
      & \multicolumn{2}{c}{TreemapMix-SCE}
      & \multicolumn{2}{c}{TreemapMix-PL} \\
    \cmidrule(lr){3-4} \cmidrule(lr){5-6} \cmidrule(lr){7-8}
    Model & Optimizer
      & Acc. $\uparrow$ & ECE $\downarrow$
      & Acc. $\uparrow$ & ECE $\downarrow$
      & Acc. $\uparrow$ & ECE $\downarrow$ \\
    \midrule
    ViT-Wee
      & AdamW  & 75.76 & 8.99 & 77.18 & \textbf{0.43} & \textbf{77.75} & 5.41 \\
      & Muon   & 77.92 & 9.88 & 78.80 & \textbf{2.04} & \textbf{79.05} & 5.93 \\
      & NAdamW & 76.30 & 9.29 & 77.48 & \textbf{0.94} & \textbf{78.16} & 5.44 \\
    \midrule
    ViT-Little
      & AdamW  & 78.77 & 8.11 & 79.50 & \textbf{1.36} & \textbf{79.93} & 5.96 \\
      & Muon   & 80.27 & 8.61 & 80.72 & \textbf{3.05} & \textbf{80.87} & 6.57 \\
      & NAdamW & 78.80 & 7.65 & 79.72 & \textbf{1.90} & \textbf{80.09} & 6.23 \\
    \midrule
    ViT-Medium
      & AdamW  & 79.89 & 8.25 & 80.33 & \textbf{0.87} & \textbf{80.71} & 5.93 \\
      & Muon   & 80.98 & 7.86 & 81.42 & \textbf{2.77} & \textbf{81.43} & 6.49 \\
      & NAdamW & 80.04 & 7.69 & 80.73 & \textbf{2.09} & \textbf{81.01} & 6.39 \\
    \midrule
    ViT-Betwixt
      & AdamW  & 81.19 & 8.11 & 81.00 & \textbf{2.02} & \textbf{81.51} & 7.05 \\
      & Muon   & 81.71 & 7.32 & \textbf{81.85} & \textbf{3.32} & 81.79 & 7.15 \\
      & NAdamW & 81.13 & 7.50 & 81.26 & \textbf{2.75} & \textbf{81.80} & 7.18 \\
    \bottomrule
  \end{tabular}
\end{table*}

\subsection{ImageNet-1K Hyperparameter Sensitivity}
\label{app:in1k-hparam-sensitivity}

TreemapMix introduces two main hyperparameters: the Dirichlet concentration
$\alpha$ and the number of mixed regions $K$. We study their effects with a
single-seed hyperparameter sweep. The sweep identifies a stable operating
regime for both TreemapMix losses: smaller $\alpha$ values and small-to-medium
values of $K$ perform best, while more extreme compositions degrade both
accuracy and calibration. Overall, TreemapMix is more sensitive to $\alpha$
than to $K$.

\begin{figure*}[!htbp]
  \centering
  \includegraphics{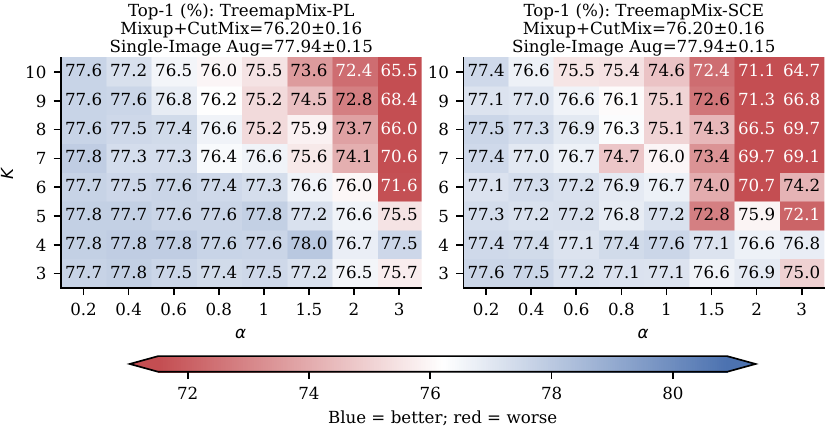}\\[8pt]
  \includegraphics{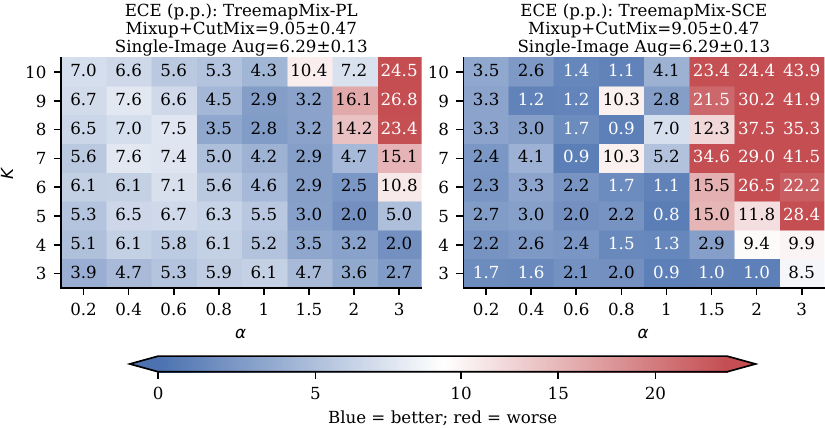}
  \caption{Appendix heatmaps for the ImageNet-1K TreemapMix hyperparameter
  sweep on ViT-Wee. ECE is measured using 15 bins. The top row shows Top-1 over the available $(K,\alpha)$ grid ($K=3,\ldots,10$),
  and the bottom row shows ECE over the same grid. The sweep uses a simplified protocol: we do not apply
  aspect filtering and we do not use $\alpha$ or $K$ scheduling, so these
  absolute values are expected to be somewhat worse than the tuned main
  results. The heatmaps show that for best results in terms of ECE and accuracy, small $K$ values together with moderately small $\alpha$ values should be used.
  }
  \label{fig:in1k-hparam-heatmaps-appendix}
\end{figure*}

The available sweep covers
$\alpha \in \{0.2, 0.4, 0.6, 0.8, 1.0, 1.5, 2.0, 3.0\}$ and
$K \in \{3,\ldots,10\}$ with the main training hyperparameters.
It uses fixed $K$ and $\alpha$ without aspect-ratio filtering.
Figure~\ref{fig:in1k-hparam-heatmaps-appendix} shows Top-1 accuracy and ECE
across this grid, with Mixup+CutMix and single-image augmentation as references.
This simplified protocol measures sensitivity; it does not reproduce the
filtered, scheduled main recipe.

For Top-1 accuracy, both TreemapMix-SCE and TreemapMix-PL perform best at
relatively small $\alpha$ values and around $K=4$. Increasing $\alpha$ up to
approximately $1$ causes only modest degradation, while larger values lead to a
clearer accuracy drop. This trend suggests that TreemapMix benefits from some
imbalance among region areas rather than from nearly uniform region sizes.
Larger $K$ values also tend to reduce Top-1 accuracy, but this effect is
weaker and less uniform than the effect of $\alpha$. As $K$ increases,
performance varies more strongly across $\alpha$ values, indicating that
high-composition settings are more sensitive to the choice of mixing strength.

Calibration, measured by ECE with 15 bins, follows a similar but sharper
pattern. For both losses, ECE remains relatively stable up to a loss-dependent
threshold and then increases substantially. For TreemapMix-SCE, this threshold
occurs around $\alpha=1$; for TreemapMix-PL, it occurs around $\alpha=1.5$.
Increasing $K$ also tends to worsen calibration, but the effect is weaker and
less systematic than the effect of $\alpha$. Across the sweep, TreemapMix-PL
is less sensitive to hyperparameter variation, while TreemapMix-SCE achieves
the strongest calibration when $\alpha$ and $K$ are chosen carefully.

These results indicate that choosing $\alpha$ is more important than choosing
$K$ for reliable TreemapMix performance. For calibration-focused use, the
sweep supports TreemapMix-SCE with $\alpha \le 1$ and $K \le 5$. For
accuracy-focused use, the sweep supports TreemapMix-PL with $\alpha \le 0.4$
and $K \le 6$. The main recipe filters layouts with extreme patch aspect ratios and schedules
$\alpha$ from $0.1$ to $0.5$. These settings were not isolated in this sweep.

\subsection{Qualitative Visualizations and Attribution Analysis}
\label{app:qualitative-visualizations}

\begin{figure*}[!htbp]
  \centering
  \setlength{\tabcolsep}{2pt}
  \begin{tabular}{cccc}
    \includegraphics{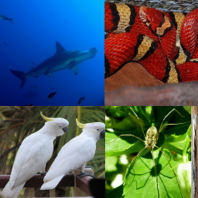} &
    \includegraphics{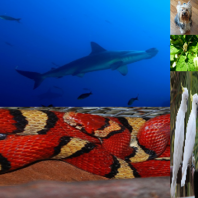} &
    \includegraphics{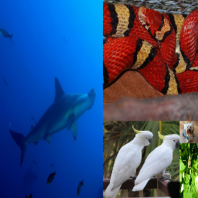} &
    \includegraphics{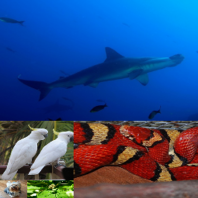} \\
    \includegraphics{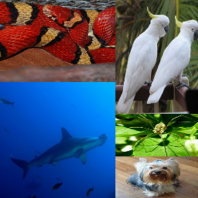} &
    \includegraphics{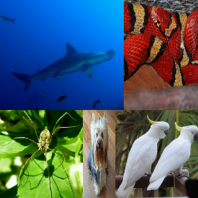} &
    \includegraphics{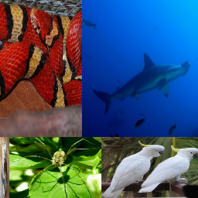} &
    \includegraphics{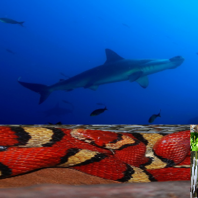}
  \end{tabular}
\caption{TreemapMix randomness for fixed sources, $\alpha=0.5$, and $K=5$. Independent Dirichlet draws produce different compositions while preserving sources and settings.}
  \label{fig:treemapmix-randomness}
\end{figure*}

\begin{figure*}[!htbp]
  \centering
  \setlength{\tabcolsep}{2pt}
  \begin{tabular}{cccc}
    \includegraphics{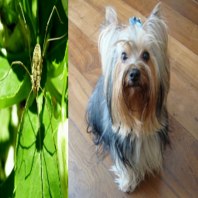} &
    \includegraphics{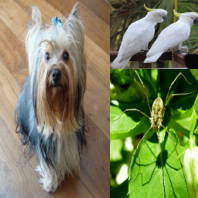} &
    \includegraphics{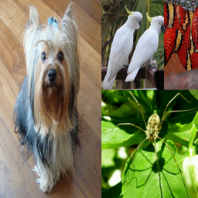} &
    \includegraphics{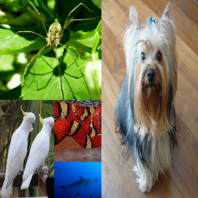} \\
    $K=2$ & $K=3$ & $K=4$ & $K=5$ \\
    \includegraphics{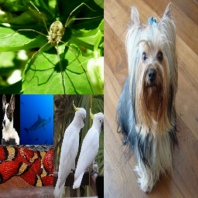} &
    \includegraphics{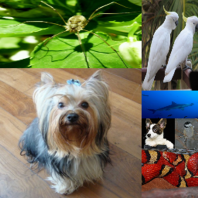} &
    \includegraphics{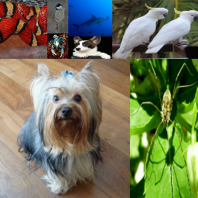} &
    \includegraphics{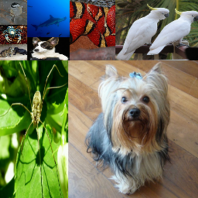} \\
    $K=6$ & $K=7$ & $K=8$ & $K=9$
  \end{tabular}
  \caption{Effect of source count for fixed images, $\alpha=2.0$, and maximum aspect ratio 15.0. Increasing $K$ from 2 to 9 gives finer-grained mixtures.}
  \label{fig:treemapmix-k-sweep}
\end{figure*}

\begin{figure*}[!p]
  \centering
  \setlength{\tabcolsep}{2pt}
  \begin{tabular}{cccc}
    \includegraphics{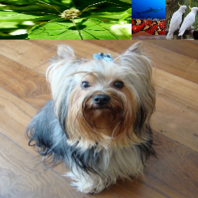} &
    \includegraphics{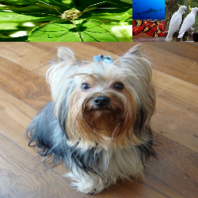} &
    \includegraphics{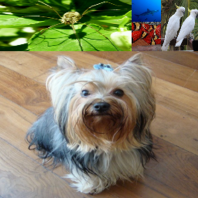} &
    \includegraphics{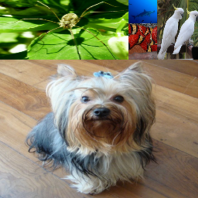} \\
    $\alpha=0.05$ & $\alpha=0.1$ & $\alpha=0.3$ & $\alpha=0.5$ \\[2pt]
    \includegraphics{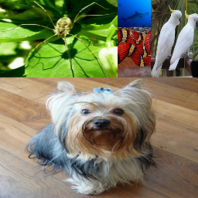} &
    \includegraphics{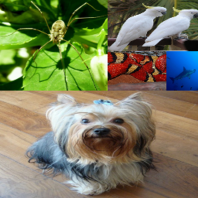} &
    \includegraphics{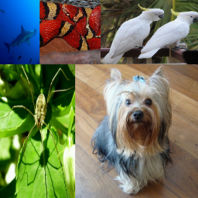} &
    \includegraphics{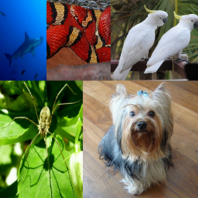} \\
    $\alpha=1.0$ & $\alpha=1.5$ & $\alpha=3.0$ & $\alpha=5.0$ \\[2pt]
  \end{tabular}
  \caption{Effect of Dirichlet concentration for fixed images, $K=5$, and maximum aspect ratio 15.0. Smaller $\alpha$ favors more unequal region areas.}
  \label{fig:treemapmix-alpha-sweep}
\end{figure*}

Figures~\ref{fig:treemapmix-randomness}, \ref{fig:treemapmix-k-sweep}, and
\ref{fig:treemapmix-alpha-sweep} visualize how TreemapMix changes with its
random draw and main hyperparameters. Figure~\ref{fig:treemapmix-randomness}
holds the source images and hyperparameters fixed while varying only the
random draw. Figures~\ref{fig:treemapmix-k-sweep} and
\ref{fig:treemapmix-alpha-sweep} sweep the number of regions and the
Dirichlet concentration, respectively, showing how these controls change the
granularity and balance of the mixed image.

\begin{figure*}[!p]
  \centering
  \includegraphics{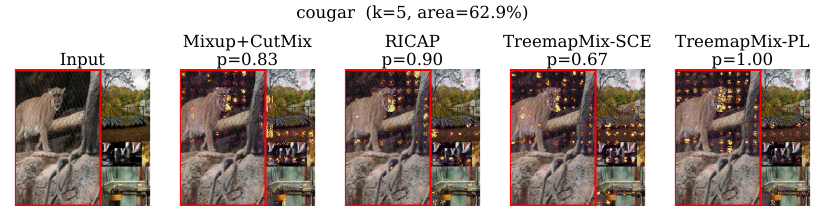}\\[3pt]
  \includegraphics{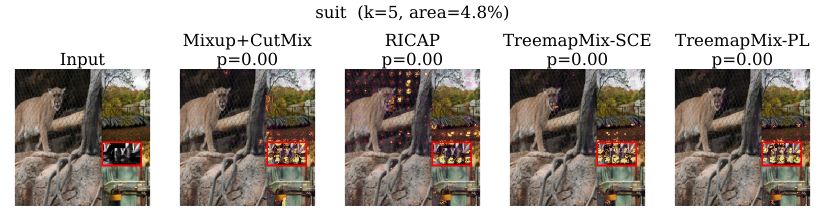}\\[3pt]
  \includegraphics{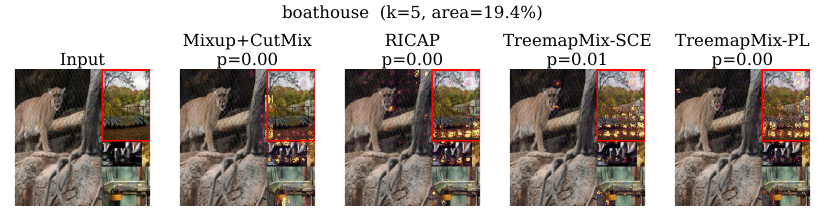}
\caption{Gradient$\times$Input attribution for three ImageNet-1K classes. Each row shows the boxed input and overlays for Mixup+CutMix, RICAP, SCE, and PL. Red boxes identify target patches; brighter regions indicate larger attribution. Labels give target probabilities and patch areas. TreemapMix often concentrates attribution on target regions; PL probabilities are sharper than SCE.}
  \label{fig:feature-space-analysis}
\end{figure*}

We also complement the quantitative logit-alignment diagnostics with
qualitative attribution maps in Figure~\ref{fig:feature-space-analysis}. The
maps are computed on the same held-out diagnostic set using the ViT-Betwixt
checkpoints. We use Gradient$\times$Input attribution
\citep{shrikumarNotJustBlack2017a,simonyanDeepConvolutionalNetworks2014},
which multiplies the input by the gradient of the target logit with respect to
the input. This provides a qualitative view of which source regions contribute
most strongly to a given class logit.

Figure~\ref{fig:feature-space-analysis} shows different probability behavior across the losses. For boathouse, SCE assigns 0.01 while other methods round to 0.00; rounded zeros need not be exact zeros. PL assigns substantially higher probability to the largest region, consistent with ranking rather than probability matching.

TreemapMix attribution often concentrates on target regions, including the small suit and boathouse patches; baseline Cougar attributions focus more on the face. These qualitative observations complement the logit diagnostics rather than providing standalone evidence. PL shows the clearest ordinal correspondence.

\clearpage
\section*{NeurIPS Paper Checklist}

\begin{enumerate}

\item {\bf Claims}
    \item[] Question: Do the main claims made in the abstract and introduction accurately reflect the paper's contributions and scope?
    \item[] Answer: \answerYes{} % Replace by \answerYes{}, \answerNo{}, or \answerNA{}.
    \item[] Justification: The abstract and introduction state the paper's main contributions and intended scope, and the later methodological, experimental, and appendix sections provide the corresponding support for these claims.

\item {\bf Limitations}
    \item[] Question: Does the paper discuss the limitations of the work performed by the authors?
    \item[] Answer: \answerYes
    \item[] Justification: We discuss limitations in Section~\ref{sec:limitations}, including area--semantic mismatch, unisolated components, computational measurement scope, long-horizon coverage, and long-tailed calibration.

\item {\bf Theory assumptions and proofs}
    \item[] Question: For each theoretical result, does the paper provide the full set of assumptions and a complete (and correct) proof?
    \item[] Answer: \answerYes{} % Replace by \answerYes{}, \answerNo{}, or \answerNA{}.
    \item[] Justification: Section~\ref{sec:method} and Appendices~\ref{app:soft-ce}, \ref{app:pl}, and \ref{app:cyclic-balancing} give the results and proofs, with distinct-label, unfiltered-sampling, and fixed-logit assumptions stated explicitly where required.

    \item {\bf Experimental result reproducibility}
    \item[] Question: Does the paper fully disclose all the information needed to reproduce the main experimental results of the paper to the extent that it affects the main claims and/or conclusions of the paper (regardless of whether the code and data are provided or not)?
    \item[] Answer: \answerYes{} % Replace by \answerYes{}, \answerNo{}, or \answerNA{}.
    \item[] Justification: The paper specifies the training protocol, augmentation settings, baselines, evaluation metrics, and seed-averaged reporting used in the main experiments, with additional hyperparameter and implementation details collected in the appendix to support reproduction of the reported results.

\item {\bf Open access to data and code}
    \item[] Question: Does the paper provide open access to the data and code, with sufficient instructions to faithfully reproduce the main experimental results, as described in supplemental material?
    \item[] Answer: \answerYes{} % Replace by \answerYes{}, \answerNo{}, or \answerNA{}.
    \item[] Justification: TreemapMix augmentation, stable supervision losses, and training instructions are available at \url{https://github.com/Ejafa/treemapmix}. Historical experiment code remains at \url{https://github.com/Ejafa/LabelMix}; Appendix~\ref{app:loss-code} distinguishes the mathematical references from historical implementation behavior.

\item {\bf Experimental setting/details}
    \item[] Question: Does the paper specify all the training and test details (e.g., data splits, hyperparameters, how they were chosen, type of optimizer) necessary to understand the results?
    \item[] Answer: \answerYes
    \item[] Justification: The paper reports the data setting, training protocols, augmentation configurations, baselines, evaluation metrics, and optimizer-related details needed to interpret the results, with appendix tables providing the fixed hyperparameters and additional implementation details.

\item {\bf Experiment statistical significance}
    \item[] Question: Does the paper report error bars suitably and correctly defined or other appropriate information about the statistical significance of the experiments?
    \item[] Answer: \answerYes
    \item[] Justification: We report means and one sample standard deviation over three seeds for the ImageNet-1K, CIFAR-100, and cross-architecture comparisons, three seeds for ImageNet-LT, and five seeds (42--46) for the loader benchmark. COCO results summarize three pretraining seeds (42--44), each with one transfer run using transfer seed 42; the optimizer and hyperparameter sensitivity studies use single-seed runs and do not estimate variation across seeds.

\item {\bf Experiments compute resources}
    \item[] Question: For each experiment, does the paper provide sufficient information on the computer resources (type of compute workers, memory, time of execution) needed to reproduce the experiments?
    \item[] Answer: \answerYes
    \item[] Justification: We explain the compute resources used for the experiments in the experiments section, including the relevant hardware setting for reproducing the reported runs.

\item {\bf Code of ethics}
    \item[] Question: Does the research conducted in the paper conform, in every respect, with the NeurIPS Code of Ethics \url{https://neurips.cc/public/EthicsGuidelines}?
    \item[] Answer: \answerYes
    \item[] Justification: We adhere to the NeurIPS Code of Ethics.

\item {\bf Broader impacts}
    \item[] Question: Does the paper discuss both potential positive societal impacts and negative societal impacts of the work performed?
    \item[] Answer: \answerNA{} % Replace by \answerYes{}, \answerNo{}, or \answerNA{}.
    \item[] Justification: We propose a general technique for image classification training, and its possible downstream uses are too broad and indirect to support a concrete societal-impact analysis within the scope of this paper.

\item {\bf Safeguards}
    \item[] Question: Does the paper describe safeguards that have been put in place for responsible release of data or models that have a high risk for misuse (e.g., pre-trained language models, image generators, or scraped datasets)?
    \item[] Answer: \answerNA{}
    \item[] Justification: The paper introduces a general image-classification training method and does not release any high-risk models, scraped datasets, or other assets that would require special misuse safeguards.

\item {\bf Licenses for existing assets}
    \item[] Question: Are the creators or original owners of assets (e.g., code, data, models), used in the paper, properly credited and are the license and terms of use explicitly mentioned and properly respected?
    \item[] Answer: \answerYes % Replace by \answerYes{}, \answerNo{}, or \answerNA{}.
    \item[] Justification: We use existing open-source software libraries and standard public datasets, cite the original sources, and follow their respective licenses and terms of use. We do not use proprietary assets

\item {\bf New assets}
    \item[] Question: Are new assets introduced in the paper well documented and is the documentation provided alongside the assets?
    \item[] Answer: \answerYes{} % Replace by \answerYes{}, \answerNo{}, or \answerNA{}.
    \item[] Justification: We release the implementation of TreemapMix, including the augmentation itself and both proposed supervision variants.

\item {\bf Crowdsourcing and research with human subjects}
    \item[] Question: For crowdsourcing experiments and research with human subjects, does the paper include the full text of instructions given to participants and screenshots, if applicable, as well as details about compensation (if any)?
    \item[] Answer: \answerNA{} % Replace by \answerYes{}, \answerNo{}, or \answerNA{}.
    \item[] Justification: The paper does not involve crowdsourcing nor research with human subjects.
    \item[] Guidelines:
    \begin{itemize}
        \item The answer \answerNA{} means that the paper does not involve crowdsourcing nor research with human subjects.
        \item Justification: Including this information in the supplemental material is fine, but if the main contribution of the paper involves human subjects, then as much detail as possible should be included in the main paper.
        \item According to the NeurIPS Code of Ethics, workers involved in data collection, curation, or other labor should be paid at least the minimum wage in the country of the data collector.
    \end{itemize}

\item {\bf Institutional review board (IRB) approvals or equivalent for research with human subjects}
    \item[] Question: Does the paper describe potential risks incurred by study participants, whether such risks were disclosed to the subjects, and whether Institutional Review Board (IRB) approvals (or an equivalent approval/review based on the requirements of your country or institution) were obtained?
    \item[] Answer: \answerNA
    \item[] Justification: The paper does not involve crowdsourcing nor research with human subjects.

\item {\bf Declaration of LLM usage}
    \item[] Question: Does the paper describe the usage of LLMs if it is an important, original, or non-standard component of the core methods in this research? Note that if the LLM is used only for writing, editing, or formatting purposes and does \emph{not} impact the core methodology, scientific rigor, or originality of the research, declaration is not required.
    %this research?
    \item[] Answer: \answerNA
    \item[] Justification: LLMs are not an important, original, or non-standard component of the core methods in this research.
    \item[] Guidelines: The core method development in this research does not involve LLMs as any important, original, or non-standard components.
    \begin{itemize}
        \item The answer \answerNA{} means that the core method development in this research does not involve LLMs as any important, original, or non-standard components.
        \item Please refer to our LLM policy in the NeurIPS handbook for what should or should not be described.
    \end{itemize}

\end{enumerate}

\end{document}